\documentclass[11pt,a4paper]{article}

 \usepackage[utf8]{inputenc}
\usepackage[T1]{fontenc}
\usepackage[a4paper,margin=1in]{geometry}
\usepackage{amsmath,amssymb}
\usepackage{graphicx}
\usepackage{booktabs}
\usepackage{array}
\usepackage{multirow}
\usepackage{float}
\usepackage{xcolor}
\usepackage{caption}
\usepackage{authblk}
\usepackage[numbers,sort&compress]{natbib}
\usepackage{hyperref}
    
\usepackage{pgfplots}
\pgfplotsset{compat=1.18}
\usepgfplotslibrary{groupplots}

\definecolor{makeblue}{HTML}{0072B2}
\definecolor{makeorange}{HTML}{E69F00}
\definecolor{makegreen}{HTML}{009E73}
\definecolor{makepurple}{HTML}{CC79A7}

\pgfplotsset{
    makeaxis/.style={
        width=0.95\linewidth,
        height=6.5cm,
        font=\footnotesize,
        tick label style={font=\footnotesize},
        label style={font=\small},
        legend style={
            font=\footnotesize,
            draw=none,
            fill=none
        },
        axis line style={black!60},
        tick style={black!60},
        ymajorgrids=true,
        grid style={black!12},
        scaled y ticks=false,
        enlarge x limits=0.15
    }
}
\usetikzlibrary{arrows.meta,positioning,fit,calc}
\usepackage{tikz}
\usetikzlibrary{arrows.meta}
\AtBeginDocument{\setlength{\headheight}{20pt}}
\title{Evaluating Aviation Safety Knowledge Extraction with
SmolLM2: From Zero-Shot to Retrieval-Grounded Prompting}

\author[1]{Aziida Nanyonga\thanks{Corresponding author:
aziida.nanyonga@kfupm.edu.sa}}
\author[2,3]{Mustafa Tahsin Yilmaz}
\author[1]{Amir Hussain}
\author[4]{Ugur Turhan}
\author[4]{Graham Wild}

\affil[1]{SDAIA-KFUPM Joint Research Center for Artificial
Intelligence, King Fahd University of Petroleum and
Minerals (KFUPM), Dhahran, Saudi Arabia}

\affil[2]{Department of Industrial Engineering,
Faculty of Engineering, King Abdulaziz University,
Jeddah 21589, Saudi Arabia}

\affil[3]{Center for Research Excellence in Artificial
Intelligence and Data Science (AIADS),
King Abdulaziz University, Jeddah 21589, Saudi Arabia}

\affil[4]{School of Science, University of New South Wales
Canberra, Australian Defence Force Academy,
Canberra, Australia}

\date{}

\begin{document}

\maketitle

\begin{abstract}
Aviation accident narratives contain safety knowledge that is difficult to capture using structured occurrence fields alone. This study evaluates prompting-based knowledge extraction across four dimensions: concept accuracy, verbatim evidence grounding, structural compliance, and directed relation extraction. SmolLM2-1.7B-Instruct was evaluated under zero-shot, one-shot, fixed few-shot, and retrieval-grounded prompting on a frozen benchmark of 300 National Transportation Safety Board narratives from 2000--2020. The reconciled reference contains 3,763 concepts across eight categories and 467 relations spanning two directional types. Evaluation combined structural checks, verbatim evidence matching, expert-reviewed one-to-one semantic matching, and paired narrative-level bootstrap confidence intervals. No prompting strategy performed best across all four dimensions. Retrieval-grounded prompting achieved the highest observed concept micro-F1 of 32.27\%, but its advantage over fixed few-shot prompting was uncertain: 0.75 percentage points (95\% CI: $-0.92$ to $2.40$). One-shot prompting achieved the highest relation F1 of 6.95\%, indicating that directed relation extraction remained the principal SmolLM2 bottleneck. Compared with fixed few-shot prompting, retrieval-grounded prompting yielded a higher proportion of verbatim evidence spans (59.45\% vs. 54.02\%) but a lower proportion of generation attempts that met all structural requirements (33.3\% vs. 42.7\%). Verbatim matching establishes source presence rather than support for the associated concept. In an exploratory comparison on a subset of 35 benchmark narratives, Qwen3-8B achieved higher concept micro-F1 than SmolLM2 under all four prompting conditions and higher relation F1 under all three demonstration-based conditions. The individual 95\% confidence intervals for these paired differences all excluded zero. Qwen3-8B achieved its highest concept micro-F1 (51.37\%) and relation F1 (34.62\%) under one-shot prompting. 
The small subset limits generalization. These findings support evaluating semantic correctness, source traceability, structural reliability, and directed relation accuracy separately.

\end{abstract}


\section{Introduction}
\label{sec:introduction}

Aviation accident investigations generate textual accounts of the circumstances, event sequences, operational conditions, contributing factors, and consequences associated with safety occurrences. These narratives contain information that may not be fully represented by predefined categorical fields, particularly when occurrences involve interactions among human, technical, operational, and environmental factors. Systematic narrative analysis can therefore complement analyses based on structured occurrence data. However, relevant concepts and their relationships are embedded in free text and must be transformed into structured representations before they can be analyzed consistently at scale.

Natural language processing (NLP) has increasingly been applied to aviation safety narratives. Topic-modeling approaches have identified latent themes and their temporal evolution in safety reports \cite{Kuhn2018,Robinson2019}. Comparative applications of latent Dirichlet allocation, non-negative matrix factorization, latent semantic analysis, probabilistic latent semantic analysis, and clustering have demonstrated the potential of NLP to characterize thematic and semantic structure in National Transportation Safety Board (NTSB) narratives \cite{Nanyonga2024}. More recent research has combined contextual language representations, topic modeling, and supervised classification to identify specific pilot failures from accident reports \cite{Macedo2026}. These approaches support corpus-level analysis and document classification, but they do not necessarily recover the individual safety concepts and directed relationships expressed within each narrative.

Information extraction addresses this complementary problem by transforming natural-language text into structured representations of entities, events, and relations \cite{Pai2024OpenIE,Xu2024LLMIE}. Large language models (LLMs) allow extraction tasks to be specified through natural-language instructions, demonstrations, and output schemas. Research on specialized scientific text has shown that LLMs can generate structured information without requiring a separately trained extraction model for every target field \cite{Dagdelen2024}. This capability offers a potential route from aviation investigation narratives to machine-readable safety knowledge while retaining distinctions that would be lost when a report is reduced to a single document-level label.

Recent aviation research has begun to investigate this approach. Retrieval-supported LLMs have been evaluated for causal identification from historical accident reports \cite{Ren2025}. Transformer-based architectures have been developed to identify cause, effect, and embedded-causality spans in NTSB narratives \cite{Chen2026ARCEM}, while other LLM-based methods have targeted accident-causation extraction and aviation safety knowledge-graph construction \cite{Chen2024CausationKG,Cheng2026}. These studies demonstrate progress toward extracting and representing safety information at a finer level than document classification.

Nevertheless, aviation safety knowledge extends beyond an isolated causal statement. A narrative may describe events, phases of flight, hazards, human actions, technical conditions, environmental circumstances, contributing factors, and consequences. Representing these elements within a common schema requires distinctions between their conceptual roles, even when they occur within the same sentence. Concept identification and relation reconstruction also impose different requirements. A model may identify two relevant concepts while assigning an incorrect relation label, reversing their direction, or substituting an upstream condition for the direct source event. Evaluation restricted to isolated concepts or aggregate causal statements may therefore overlook errors that affect the reliability of the resulting knowledge representation.

Grounding introduces a further challenge. An LLM may produce an explanation that is plausible in aviation terms but unsupported by the particular investigation narrative. Such an explanation should not be treated as a reported finding solely because it is technically reasonable. Extraction accuracy must therefore be distinguished from source traceability: a concept can correspond semantically to a reference annotation even when its accompanying quotation is inaccurate, and a quotation can occur verbatim in the narrative without supporting the concept assigned to it. Checking verbatim evidence presence provides an auditable measure of traceability, but it does not establish evidential relevance.

Structural reliability is another requirement for downstream use. A response may be parseable as JSON while omitting required categories, violating the evidence structure, or using relation labels outside the permitted vocabulary. Conversely, a structurally compliant response may contain semantically incorrect concepts or relations. Semantic accuracy, evidence traceability, structural compliance, and directed relation correctness should consequently be evaluated as separate dimensions.

The contextual guidance supplied through prompting may affect these dimensions differently. Zero-shot prompting provides instructions without worked examples, whereas one-shot and few-shot prompting introduce demonstrations of category boundaries, extraction granularity, and output structure. Retrieval-grounded prompting dynamically selects demonstrations according to their similarity to the target narrative. In this study, retrieval supplies annotated development examples rather than external accident evidence; the target narrative remains the source from which knowledge must be extracted. Although retrieval-supported methods have shown promise for aviation causation analysis \cite{Ren2025}, the relative performance of fixed and dynamically selected demonstrations remains insufficiently characterized for multidimensional extraction under a common model, schema, benchmark, and evaluation protocol.

The present study evaluates structured aviation safety knowledge extraction using the compact instruction-tuned SmolLM2-1.7B-Instruct model. The extraction schema comprises eight concept categories: event, phase of flight, hazard, human factor, technical factor, environmental factor, contributing factor, and consequence. Two directional relation types, \texttt{CONTRIBUTES\_TO} and \texttt{RESULTS\_IN}, represent relationships among these concepts. The output specification also requires textual evidence, allowing concept accuracy and verbatim source traceability to be assessed independently.

Four prompting conditions are compared: zero-shot, one-shot, fixed few-shot, and retrieval-grounded prompting. The experiment uses the same frozen set of 300 NTSB narratives from 2000--2020 and a completed reconciled reference containing 3,763 concepts and 467 directed relations. One-shot prompting supplies one fixed annotated development example, while fixed few-shot and retrieval-grounded prompting each supply four examples. Retrieval selects its examples exclusively from a frozen development pool that is separate from the evaluation benchmark. No task-specific fine-tuning is performed, and the model checkpoint, generation configuration, extraction schema, and evaluation narratives remain unchanged across conditions.

Concept extraction is evaluated through candidate-assisted, expert-reviewed one-to-one semantic matching within each narrative and category. Relation evaluation additionally requires correspondence of the source, target, exact relation label, and direction. Structural compliance and verbatim evidence grounding are assessed separately from these semantic measures. Category-specific and narrative-length analyses characterize variation in performance, while paired narrative-level bootstrap confidence intervals quantify sampling uncertainty in estimates and differences between prompting conditions.

An exploratory supplementary experiment compares SmolLM2-1.7B-Instruct with Qwen3-8B on the same 35 benchmark narratives under all four prompting conditions. Both models are scored against the corresponding reconciled reference containing 432 concepts and 77 relations, and SmolLM2 results are recomputed from its existing outputs for this subset. The comparison therefore controls for the evaluated narratives and reference annotations. However, the subset was selected by processing order under computational constraints rather than random sampling, and differences between the models extend beyond parameter count. The experiment provides exploratory evidence about their relative behavior rather than a full-benchmark ranking or an isolated test of model size.

The study addresses the following research questions:

\begin{itemize}
    \item \textbf{RQ1:} How accurately can a compact instruction-tuned language model extract structured aviation safety concepts and directed relationships from accident investigation narratives?

    \item \textbf{RQ2:} How do zero-shot, one-shot, fixed few-shot, and retrieval-grounded prompting affect semantic extraction performance and structured-output reliability?

    \item \textbf{RQ3:} What proportion of generated evidence spans can be traced verbatim to the source narratives, and how does this measure vary across prompting conditions?

    \item \textbf{RQ4:} How does extraction performance vary across safety-knowledge categories and narrative-length strata, and how does concept identification compare with directed relation extraction?
\end{itemize}

The study contributes a controlled comparison of four prompting strategies across four complementary evaluation dimensions, using a common extraction schema and reconciled reference. It also provides category-specific and narrative-length analyses, paired uncertainty estimates, and an exploratory comparison with a larger instruction-tuned model. The distinction between verbatim traceability and evidential support is maintained throughout: locating a generated quotation in the source narrative does not establish that it justifies the associated concept.

Two principal findings emerge from the main SmolLM2 experiment. First, no prompting strategy performs best across all evaluation dimensions. Retrieval-grounded prompting achieves the highest observed concept micro-F1 of 32.27\%, but its paired difference from fixed few-shot prompting is uncertain at 0.75 percentage points (95\% CI: $-0.92$ to $2.40$). Retrieval also produces a higher verbatim-grounding rate, whereas one-shot prompting achieves the highest complete structural compliance. Second, directed relation extraction remains the principal SmolLM2 bottleneck, with a highest relation F1 of 6.95\% under one-shot prompting. The matched Qwen3-8B comparison shows higher concept F1 across all conditions and higher relation F1 under demonstration-based prompting, although its small, non-randomly selected subset restricts generalization.

The remainder of this paper is organized as follows. Section~2 reviews related work on aviation narrative analysis, information and causality extraction, structured safety knowledge, prompting, and evidence grounding. Section~3 describes the dataset, extraction schema, reference construction, inference configuration, prompting conditions, and evaluation procedures. Section~4 presents the results, including category-specific and narrative-length analyses and the matched Qwen3-8B comparison. Section~5 discusses the findings, implications, and limitations. Section~6 concludes the paper and identifies directions for future research.
\section{Related Work}
\label{sec:related-work}

\subsection{NLP for Aviation Safety Narratives}

Aviation safety databases contain narrative reports describing
operational events, hazards, human actions, technical conditions,
and environmental circumstances. NLP provides methods for analyzing
these reports systematically and identifying patterns that may be
difficult to detect through manual review alone. Topic-modeling
approaches have been used to identify latent themes and temporal
changes in aviation safety reports \cite{Kuhn2018,Robinson2019},
supporting scalable exploration of recurring safety issues.

Subsequent studies have examined alternative topic-modeling,
clustering, and representation-learning techniques for organizing
accident and incident narratives
\cite{Nanyonga2024,Nanyonga2025,Nanyonga2025Socrata}.
Supervised NLP and deep learning approaches have also been
investigated for predicting injury levels and classifying aviation
safety incidents using Australian occurrence narratives
\cite{Nanyonga2025Injury,Nanyonga2025Incident}.
These studies demonstrate how textual analysis can complement
structured occurrence variables. More recent research has combined
Bidirectional Encoder Representations from Transformers (BERT)
representations with topic modeling and supervised classification
to identify specific pilot-failure categories in NTSB accident
reports \cite{Macedo2026}. Related work has integrated narrative
analysis with knowledge-graph methods to examine relationships
among aviation safety factors \cite{Farzadnia2026}.

These studies support corpus exploration, trend identification,
safety outcome prediction, and document classification. However,
assigning a topic or class to a report does not necessarily
identify the individual safety concepts expressed within it or
reconstruct the directed relationships connecting those concepts.
Structured knowledge extraction therefore requires evaluation
at a finer level than the document or document cluster.

\subsection{Information and Causality Extraction from Aviation Reports}

Information extraction transforms unstructured text into structured representations of entities, events, attributes, and relationships. Pretrained language models and LLMs have extended the available methods by allowing extraction tasks to be specified through instructions, demonstrations, and output schemas \cite{Pai2024OpenIE,Xu2024LLMIE}. Applications to specialized scientific text illustrate the potential of LLMs to extract structured information without developing a separate supervised architecture for every target field \cite{Dagdelen2024}.

Research outside aviation also provides methodological context for mapping textual expressions to predefined categories. Khatoon et al.~\cite{Khatoon2026NER} used feature-enhanced recurrent sequence labeling with conditional random field decoding to identify aspects and sentiment expressions in reviews. Mir et al.~\cite{Mir2022ImplicitAspects} investigated the mapping of implicit expressions to explicit aspect categories. These studies address fine-grained concept recognition and category assignment, but they do not directly evaluate aviation safety concepts, directed causal relations, or evidence support. Their relevance to the present study concerns the distinction between recognizing a textual expression and assigning it the appropriate conceptual role.

Within aviation, Ren et al.~\cite{Ren2025} proposed a retrieval-supported causal-identification framework that combines retrieval, noise filtering, and LLM-based generation to extract causation information from historical accident reports. Their work illustrates how retrieved context can support causal information extraction from unstructured safety data.

Chen et al.~\cite{Chen2026ARCEM} developed the Aviation Reports Causality Extraction Model (ARCEM) for NTSB narratives. ARCEM formulates causality extraction as a sequence-labeling problem and identifies cause, effect, and embedded-causality spans using contextual BERT representations within a specialized neural architecture. This formulation targets the localization of causal expressions, whereas generative extraction additionally requires the model to produce concepts and relations in the requested output structure.

Other research has used LLMs to extract aviation accident-causation concepts and relationships for subsequent knowledge-graph construction \cite{Chen2024CausationKG}. Xiang et al.~\cite{Xiang2026} combined retrieval-supported generation and prompt engineering to construct a causal knowledge graph from 343 NTSB accident reports. On a 50-report validation subset, they reported causation accuracy of 95.2--95.7\% and relation accuracy of 98.3--100\% across three language models, measured through verification of generated items against source text. These measures characterize the correctness of generated items, whereas reference-based recall and F1 additionally penalize missed annotations. Their pipeline also uses factual narratives, analysis narratives, and probable-cause statements, together with output verification and refinement. The present experiment uses only analysis narratives and preserves generated extraction content without repair. Differences in inputs, models, processing, and metric definitions therefore prevent direct comparison of the reported scores.

Causality-focused extraction addresses an important component of aviation safety analysis, but an investigation narrative may contain several additional forms of knowledge. Events, phases of flight, hazards, human actions, technical conditions, environmental circumstances, contributing factors, and consequences may all require representation. Evaluating these categories within a common schema makes it possible to examine whether a model distinguishes their roles rather than identifying only a causal statement or a limited set of cause--effect spans.

Our related study on decomposed relation extraction evaluates relation classification when reference concept endpoints are supplied, using support vector machine and RoBERTa classifiers under exhaustive ordered-pair evaluation \cite{NanyongaDecomposed}. The studies share the 300-narrative NTSB benchmark and overlapping SmolLM2 end-to-end evaluation material. They therefore do not constitute independent evaluations on separate datasets. The decomposed study examines relation classification under sparse positive relations among supplied concept pairs, whereas the present study compares prompting strategies for joint concept and relation extraction and separately evaluates structure and verbatim evidence traceability. The present manuscript reports the completed reconciled-reference evaluation; reference versions and associated scores must be distinguished when comparing overlapping experimental material.

\subsection{Knowledge Graphs and Structured Aviation Safety Knowledge}

Knowledge graphs represent concepts together with the relationships connecting them. They provide a useful representation for aviation safety analysis because occurrences may involve interactions among operational, human, technical, and environmental factors rather than isolated events.

Recent studies have integrated LLM-based extraction with aviation knowledge-graph construction. Accident-causation concepts and relations have been extracted from narratives and organized into structured causal representations \cite{Chen2024CausationKG}. Cheng et al.~\cite{Cheng2026} proposed an LLM-powered framework for constructing an aviation safety hazard causal knowledge graph, combining few-shot extraction with relation normalization and retrieval-supported reasoning. Such approaches illustrate how narrative-derived representations can support subsequent retrieval and safety analysis.

The reliability of these downstream applications depends on the extraction stage. An incorrect concept, unsupported relation, or reversed edge can propagate into the graph and affect later retrieval or reasoning. Structural validity alone is insufficient: a relation can satisfy the required schema while connecting the wrong endpoints, and a correctly identified concept can still be assigned an inappropriate category. Evaluation should therefore distinguish the correctness of extracted knowledge from the successful production of a machine-readable representation.

The present study concentrates on this extraction stage rather than evaluating a downstream graph-reasoning system. Its schema provides a structured representation of safety concepts and directed relations, while separate semantic, structural, and evidence-traceability measures characterize different sources of error before such outputs are used in subsequent analysis.

\subsection{In-Context Learning and Retrieval-Grounded Extraction}

In-context learning allows LLMs to perform a task using instructions and demonstrations supplied within the prompt, without updating model parameters. Zero-shot prompting provides instructions without demonstrations, whereas one-shot and few-shot prompting provide one or several worked examples. Demonstrations can illustrate category boundaries, extraction granularity, relation structure, and the expected output format.

Demonstration effectiveness may depend on both the number of examples and their relevance to the target input. Retrieval-augmented generation (RAG) addresses the broader problem of supplying relevant context before generation \cite{Lewis2020}. In extraction tasks, a related approach retrieves annotated examples and presents them as dynamically selected demonstrations, allowing the prompt context to vary with the target narrative.

The retrieval-grounded condition in the present study follows this demonstration-selection formulation. It retrieves annotated development examples rather than external factual evidence about the accident under analysis. The examples illustrate how narratives map to the extraction schema, while the target narrative remains the source of the concepts and evidence to be extracted. This distinction separates dynamic demonstration selection from RAG pipelines that retrieve external documents primarily to supply factual content.

Retrieval-supported methods have already been investigated for aviation causal identification and knowledge-graph reasoning \cite{Ren2025,Cheng2026}. However, their performance does not establish that dynamically selected demonstrations will improve every aspect of structured extraction. Similarity-based retrieval may select examples that resemble the target narrative without adequately covering its required concept categories or relation patterns. A controlled comparison should therefore hold the model, schema, generation configuration, and evaluation narratives constant while examining semantic accuracy, structural compliance, source traceability, and relation correctness separately.

\subsection{Evidence Grounding and Reliability of Generative Information Extraction}

Generative extraction can produce information that is plausible within a domain but unsupported by the particular source document. This distinction matters in safety analysis because pretrained knowledge about a possible failure mechanism should not be treated as a finding reported in an investigation narrative.

Linking extracted concepts to quoted passages improves auditability by allowing a reviewer to inspect the text presented as support. Nevertheless, evidence presence and evidence relevance are different properties. Verifying that a quotation occurs verbatim in a narrative establishes traceability; determining whether it supports a concept, category assignment, or causal interpretation requires an additional assessment.

Research on attributed generation similarly distinguishes answer correctness from the quality of supporting citations. Gao et al.~\cite{Gao2023ALCE} introduced a benchmark that evaluates generated answers along separate dimensions, including correctness and citation quality. This distinction is relevant to extraction because a source passage can be correctly located while providing insufficient support for the information attributed to it.

Evidence traceability also does not guarantee structural or relational correctness. A model may quote an appropriate passage while assigning its content to the wrong category, or it may extract valid concepts while reversing a relation between them. Conversely, a response may satisfy the output schema while containing unsupported information. These possibilities motivate separate evaluation of concept accuracy, verbatim evidence grounding, structural compliance, and directed relation correctness. In the present study, the grounding measure is explicitly restricted to verbatim source presence and is not interpreted as a measure of evidential support.

\subsection{Research Gap}

Existing aviation NLP research demonstrates the value of topic modeling, document classification, causal extraction, and knowledge-graph construction for analyzing safety narratives. Recent LLM-based and retrieval-supported approaches further illustrate the potential for extracting causal information and organizing it into structured representations \cite{Ren2025,Chen2024CausationKG,Chen2026ARCEM,Cheng2026,Xiang2026}. These developments motivate a complementary evaluation of prompting-based extraction under controlled experimental conditions.

Four issues define the focus of the present study. First, extraction should be assessed across multiple safety-knowledge categories within a common schema, rather than only through isolated entities or causal statements. Second, fixed and dynamically retrieved demonstrations should be compared while holding the model, schema, generation configuration, and benchmark constant. Third, semantic correctness, verbatim source traceability, structural compliance, and directed relation accuracy should be evaluated separately because improvements in one dimension need not imply improvements in another. Fourth, category-specific and narrative-length analyses are needed to characterize variation that aggregate performance measures may conceal.

The present study addresses these issues through a controlled evaluation of SmolLM2-1.7B-Instruct under zero-shot, one-shot, fixed few-shot, and retrieval-grounded prompting. All conditions use the same 300 narratives and completed reconciled reference, covering eight concept categories and two directional relation types. One-to-one semantic matching measures concept and relation accuracy, while independent evaluation procedures assess output structure and verbatim evidence presence. Category-specific and narrative-length analyses complement aggregate metrics, and paired bootstrap intervals quantify sampling uncertainty.

A supplementary comparison evaluates Qwen3-8B and SmolLM2 on the same 35 narratives against the corresponding reconciled reference. This comparison provides exploratory evidence about differences between the evaluated models while preserving the distinction between the matched subset and the full benchmark. Together, these analyses characterize the capabilities and limitations of structured generative extraction without treating source-text presence, schema compliance, or a small-subset model comparison as sufficient evidence of reliable aviation safety knowledge extraction.

\section{Materials and Methods}
\label{sec:materials-methods}

\subsection{Study Design}

This study evaluated structured aviation safety knowledge extraction using SmolLM2-1.7B-Instruct under four prompting conditions: zero-shot, one-shot, fixed few-shot, and retrieval-grounded prompting. The model checkpoint, extraction schema, generation configuration, and evaluation narratives were held constant, while the number and selection of annotated demonstrations varied. No task-specific fine-tuning or parameter updating was performed.

The workflow comprised corpus preparation, schema development, construction of development examples and a held-out benchmark, generation under the four prompting conditions, structural and verbatim-grounding evaluation, and one-to-one semantic evaluation of concepts and relations. Schema-development cases and demonstration records were excluded from the evaluation benchmark. In the retrieval-grounded condition, each target narrative served only as a query for selecting demonstrations from the frozen development pool; its reference annotations were unavailable to both the retrieval mechanism and the model.

All reported semantic results use the completed reconciled reference. Rescoring used the existing model generations without rerunning inference or modifying their content. Structural compliance and verbatim evidence-grounding measurements were evaluated separately because they depend on the saved outputs and source narratives rather than on the reference annotation version.

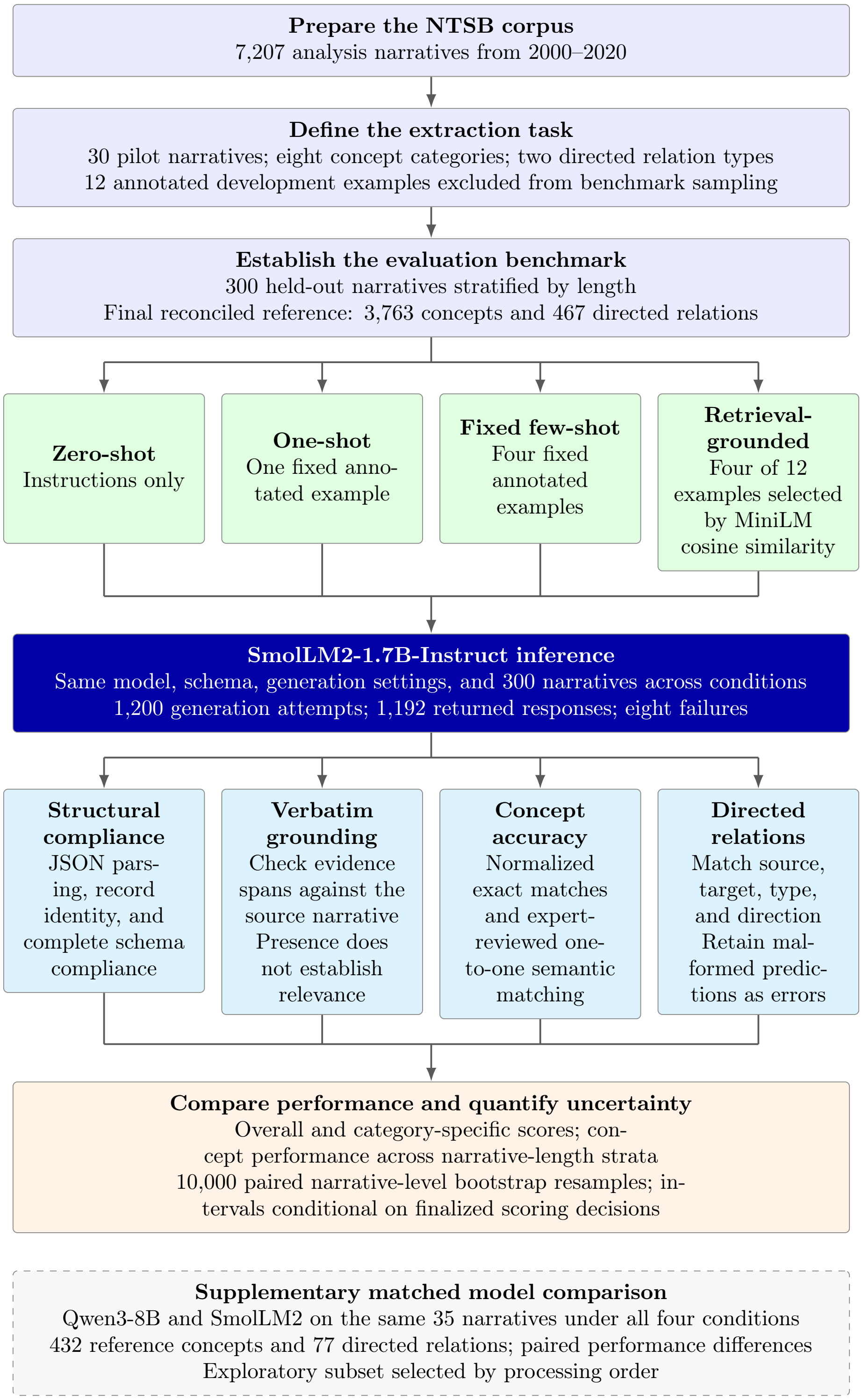
\begin{figure}[p]
\centering
\resizebox{!}{0.80\textheight}{%
\begin{tikzpicture}[
    font=\small,
    box/.style={
        draw=black!45,
        rounded corners=2pt,
        align=center,
        inner sep=6pt
    },
    preparation/.style={
        box,
        fill=blue!8,
        text width=13cm
    },
    condition/.style={
        box,
        fill=green!12,
        text width=2.8cm,
        minimum height=2.4cm
    },
    model/.style={
        box,
        fill=blue!65!black,
        text=white,
        text width=13cm
    },
    evaluation/.style={
        box,
        fill=cyan!12,
        text width=2.8cm,
        minimum height=3cm
    },
    analysis/.style={
        box,
        fill=orange!10,
        text width=13cm
    },
    supplementary/.style={
        box,
        dashed,
        fill=black!3,
        text width=13cm
    },
    flow/.style={
        -{Latex[length=2.2mm]},
        thick,
        draw=black!65
    },
    connector/.style={
        thick,
        draw=black!65
    },
    rowframe/.style={
        inner sep=0pt,
        outer sep=0pt,
        draw=none,
        fill=none
    }
]

\node[preparation] (corpus) {
    \textbf{Prepare the NTSB corpus}\\
    7,207 analysis narratives from 2000--2020
};

\node[preparation,below=5mm of corpus] (schema) {
    \textbf{Define the extraction task}\\
    30 pilot narratives; eight concept categories;
    two directed relation types\\
    12 annotated development examples excluded
    from benchmark sampling
};

\node[preparation,below=5mm of schema] (reference) {
    \textbf{Establish the evaluation benchmark}\\
    300 held-out narratives stratified by length\\
    Final reconciled reference:
    3,763 concepts and 467 directed relations
};

\draw[flow] (corpus.south) -- (schema.north);
\draw[flow] (schema.south) -- (reference.north);

\coordinate (prompttop)
    at ($(reference.south)+(0,-9mm)$);

\node[condition,anchor=north] (zero)
    at ($(prompttop)+(-5.25,0)$) {
    \textbf{Zero-shot}\\
    Instructions only
};

\node[condition,anchor=north] (one)
    at ($(prompttop)+(-1.75,0)$) {
    \textbf{One-shot}\\
    One fixed annotated example
};

\node[condition,anchor=north] (few)
    at ($(prompttop)+(1.75,0)$) {
    \textbf{Fixed few-shot}\\
    Four fixed annotated examples
};

\node[condition,anchor=north] (retrieval)
    at ($(prompttop)+(5.25,0)$) {
    \textbf{Retrieval-grounded}\\
    Four of 12 examples selected by
    MiniLM cosine similarity
};

\node[rowframe,fit=(zero)(one)(few)(retrieval)]
    (promptrow) {};

\coordinate (promptbus)
    at ($(reference.south)+(0,-4mm)$);

\draw[connector] (reference.south) -- (promptbus);
\draw[connector]
    (zero.north |- promptbus)
    -- (retrieval.north |- promptbus);

\foreach \name in {zero,one,few,retrieval}{
    \draw[flow]
        (\name.north |- promptbus)
        -- (\name.north);
}

\node[model,below=10mm of promptrow] (generation) {
    \textbf{SmolLM2-1.7B-Instruct inference}\\
    Same model, schema, generation settings,
    and 300 narratives across conditions\\
    1,200 generation attempts;
    1,192 returned responses; eight failures
};

\coordinate (mergebus)
    at ($(promptrow.south)+(0,-4mm)$);

\foreach \name in {zero,one,few,retrieval}{
    \draw[connector]
        (\name.south)
        -- (\name.south |- mergebus);
}

\draw[connector]
    (zero.south |- mergebus)
    -- (retrieval.south |- mergebus);
\draw[flow] (mergebus) -- (generation.north);

\coordinate (evaluationtop)
    at ($(generation.south)+(0,-9mm)$);

\node[evaluation,anchor=north] (structure)
    at ($(evaluationtop)+(-5.25,0)$) {
    \textbf{Structural compliance}\\
    JSON parsing, record identity,
    and complete schema compliance
};

\node[evaluation,anchor=north] (evidence)
    at ($(evaluationtop)+(-1.75,0)$) {
    \textbf{Verbatim grounding}\\
    Check evidence spans against
    the source narrative\\
    Presence does not establish relevance
};

\node[evaluation,anchor=north] (concepts)
    at ($(evaluationtop)+(1.75,0)$) {
    \textbf{Concept accuracy}\\
    Normalized exact matches and
    expert-reviewed one-to-one
    semantic matching
};

\node[evaluation,anchor=north] (relations)
    at ($(evaluationtop)+(5.25,0)$) {
    \textbf{Directed relations}\\
    Match source, target, type,
    and direction\\
    Retain malformed predictions as errors
};

\node[rowframe,fit=(structure)(evidence)(concepts)(relations)]
    (evaluationrow) {};

\coordinate (evaluationbus)
    at ($(generation.south)+(0,-4mm)$);

\draw[connector]
    (generation.south) -- (evaluationbus);
\draw[connector]
    (structure.north |- evaluationbus)
    -- (relations.north |- evaluationbus);

\foreach \name in {structure,evidence,concepts,relations}{
    \draw[flow]
        (\name.north |- evaluationbus)
        -- (\name.north);
}

\node[analysis,below=10mm of evaluationrow] (analysis) {
    \textbf{Compare performance and quantify uncertainty}\\
    Overall and category-specific scores;
    concept performance across narrative-length strata\\
    10,000 paired narrative-level bootstrap resamples;
    intervals conditional on finalized scoring decisions
};

\coordinate (analysisbus)
    at ($(evaluationrow.south)+(0,-4mm)$);

\foreach \name in {structure,evidence,concepts,relations}{
    \draw[connector]
        (\name.south)
        -- (\name.south |- analysisbus);
}

\draw[connector]
    (structure.south |- analysisbus)
    -- (relations.south |- analysisbus);
\draw[flow] (analysisbus) -- (analysis.north);

\node[supplementary,below=6mm of analysis] (qwen) {
    \textbf{Supplementary matched model comparison}\\
    Qwen3-8B and SmolLM2 on the same
    35 narratives under all four conditions\\
    432 reference concepts and 77 directed relations;
    paired performance differences\\
    Exploratory subset selected by processing order
};

\end{tikzpicture}%
}
\caption{Experimental framework for aviation safety knowledge
extraction, showing corpus preparation, the extraction schema,
four prompting conditions, model inference, and four separate
evaluation dimensions. The dashed box summarizes the
supplementary matched model comparison.}
\label{fig:overall_framework}
\end{figure}

Figure~\ref{fig:overall_framework} summarizes the experimental framework.

\subsection{NTSB Aviation Accident Dataset}

The study used NTSB records covering 2000--2020. The available data included structured occurrence variables and textual fields describing accident circumstances and investigative findings. Principal fields included \texttt{HighestInjury}, \texttt{damageLevel}, \texttt{aircraftCategory}, \texttt{flightOperationType}, \texttt{eventType}, \texttt{probableCause}, and \texttt{analysisNarrative}.

The \texttt{analysisNarrative} field was the sole textual input to the models and the source used for reference annotation. The \texttt{probableCause} field was withheld from prompting, retrieval, and annotation to prevent extraction from relying on the separate probable-cause statement. Structured occurrence metadata were retained for corpus characterization but were not supplied to the models.

Initial preprocessing identified 7,208 usable narrative records. One record contained only an eight-word pointer to external Web content rather than a substantive investigation narrative. Its exclusion produced a corpus of 7,207 unique substantive narratives.

Narrative length was measured by word count. Narratives containing at most 92 words were classified as short, those containing 93--195 words as medium, and those containing more than 195 words as long. The corpus contained 1,824 short, 3,589 medium, and 1,794 long narratives. These strata were used to represent concise and detailed reports during development and benchmark sampling.

\subsection{Aviation Safety Knowledge Extraction Schema}

The extraction schema comprised eight categories:

\begin{itemize}
    \item \textbf{Event}: an occurrence, action, or operational development described as having taken place;
    \item \textbf{Phase of flight}: the operational phase during which the relevant occurrence took place;
    \item \textbf{Hazard}: a condition, situation, or circumstance presenting a safety threat;
    \item \textbf{Human factor}: a relevant human action, decision, omission, performance limitation, or other explicitly described human-related factor;
    \item \textbf{Technical factor}: an explicitly described aircraft, system, component, equipment, or mechanical factor;
    \item \textbf{Environmental factor}: a relevant weather, terrain, visibility, atmospheric, or other environmental condition;
    \item \textbf{Contributing factor}: an explicitly supported contributory circumstance not more appropriately represented by another category; and
    \item \textbf{Consequence}: an outcome of an occurrence, including injury, damage, impact, loss of control, or another explicitly described result.
\end{itemize}

Two directional relation types represented connections between concepts:

\begin{itemize}
    \item \texttt{CONTRIBUTES\_TO}: the source concept explicitly contributes to the target concept; and
    \item \texttt{RESULTS\_IN}: the source concept explicitly produces or leads to the target event or consequence.
\end{itemize}

Category assignment depended on the role of the information within the narrative. An action was not automatically a human factor, a component mention was not automatically a technical factor, and an impact description was not automatically interchangeable with the obstacle or terrain hazard involved. Similarly, identifying two related concepts did not establish that a directed relation existed between them.

Three coauthors serving as aviation-domain reviewers annotated the benchmark using the common schema and evidence-based codebook. Their individual submissions were retained separately from subsequent reconciliation. Differences in inclusion, category assignment, granularity, evidence selection, and relation specification were discussed, and final annotations were recorded with supporting evidence. Case-identification and transcription discrepancies were resolved without treating wrong-case submissions as valid independent judgments for the intended narrative.

\subsection{Annotation Protocol and Reference Standard}

Reference annotation followed a conservative evidence-based protocol. Each concept and relation had to be supported by the \texttt{analysisNarrative}. Annotators were not permitted to introduce information from \texttt{probableCause}, structured metadata, external aviation knowledge, or assumptions about what would normally occur in similar circumstances.

Evidence spans were recorded as verbatim substrings. Textual presence was distinguished from causal or contributory interpretation: mentioning a condition did not establish that it contributed to the occurrence, and temporal precedence alone did not establish a directed relation. Corrective actions were not treated as human errors unless the narrative supported that interpretation. Recommendations and counterfactual statements were not annotated as events that occurred. Negated technical problems were excluded as present technical factors, and accident-induced damage was distinguished from pre-existing technical conditions. Weather statements were annotated according to their relevance to the occurrence rather than automatically treated as safety factors.

The completed reconciled reference contains 3,763 concepts and 467 directed relations across the 300 benchmark narratives. An earlier reference containing 2,990 concepts and 248 relations was retained for provenance, but its semantic scores are superseded by the reconciled evaluation reported here. Reconciliation included additions, wording and granularity changes, and corrections; consequently, the reference versions were not treated as interchangeable scoring resources.

Existing matching decisions were reused only when their narrative context, category, and prediction--reference content remained applicable. Additional or changed matching problems were reviewed separately. The final scoring reference and completed matching decisions were then fixed before calculating the reported metrics and intervals.

\subsubsection{Schema-Development Pilot}

A 30-case pilot was used exclusively for schema development and annotation-rule refinement. It contained 24 accident records and six incident records, with ten short, ten medium, and ten long narratives.

The pilot supported refinement of category boundaries, evidence requirements, and relation rules. The schema and guidance were subsequently frozen before final model evaluation. Pilot cases were excluded from the evaluation benchmark.

\subsubsection{Development Examples}

Following schema stabilization, 12 annotated development examples were prepared as in-context demonstrations. These examples were frozen before inference and excluded from the evaluation benchmark. They formed the complete candidate pool for both fixed and retrieval-selected demonstrations.

The examples illustrated category assignment, extraction granularity, evidence representation, relation structure, and the required output format. They were not used for parameter optimization or task-specific training.

\subsubsection{Final Evaluation Benchmark}

After excluding the 30 pilot cases and 12 development examples, 7,165 narratives remained eligible for benchmark sampling. Stratified random sampling with seed 42 selected 300 narratives: 76 short, 150 medium, and 74 long.

The selected narratives had a mean length of 164.31 words, a standard deviation of 112.47 words, a median of 136 words, and a range of 34--1,013 words. Narrative membership remained unchanged across prompting conditions and reference reconciliation.

The final reference contains 1,285 events, 613 phases of flight, 536 hazards, 212 human factors, 154 technical factors, 394 environmental factors, nine contributing factors, and 560 consequences, totaling 3,763 concepts. It additionally contains 467 directed relations.

The concept-evidence fields contain 3,775 recorded span entries, verified as verbatim substrings after transcription and formatting corrections. These entries are not necessarily unique passages: multiple spans can support one concept, and a shared span can support multiple concepts. Semantic scoring uses concept and relation instances as its denominators rather than the number of evidence entries.

\subsubsection{Agreement Assessment}
\label{sec:annotation-agreement}

Reference-annotation agreement and model-to-reference matching agreement concern different judgments. Reference diagnostics examine consistency among the retained annotation submissions, whereas matching decisions determine whether a model prediction corresponds to a reference item. Final consensus approval was not treated as an independent agreement observation.

Reference-annotation diagnostics included 232 narratives with parseable lists, matching case identifiers, and a single unambiguous Reviewer~3 submission. The remaining 68 narratives were excluded because 57 had Reviewer~3 case-ID mismatches, ten had multiple submitted versions, and one had an invalid submission. Subsequent reconciliation did not recover independent pre-discussion judgments for these excluded cases.

For each eligible narrative and category, a nonempty annotation list was coded as presence and an empty list as absence. Three-reviewer unanimity and category-specific Fleiss' $\kappa$ were calculated. Percentile 95\% confidence intervals used 10,000 narrative-level bootstrap resamples with seed 42. Relation presence was evaluated separately using the same procedure.

These diagnostics assess agreement on whether a category contains an annotation, not agreement on the identity, completeness, or evidential support of individual concepts or on complete directed relations. Their restriction to 232 eligible narratives also limits interpretation across the full benchmark.

All three reviewers confirmed the final reconciled annotations for the 300 narratives. These confirmations document approval of the reference and were retained separately from the submissions used for agreement diagnostics.

The semantic-matching records combine historical adjudications, reused decisions, supplied reviewer pair sets, and explicit final corrections. Although available pair sets were preserved, the consolidated records did not establish independent pre-discussion provenance for the entire reconciled matching pass. No new independent matching-agreement coefficient was therefore inferred from these fields. Historical supplementary agreement estimates from the earlier-reference evaluation were not treated as agreement measurements for the expanded reconciled evaluation.

\subsection{Structured Model Output}

The models were instructed to return a JSON object containing the supplied record identifier, a list of strings for each concept category, an evidence object containing category-specific lists of quoted strings, and a list of directed relations. The record identifier originated in the dataset metadata, was supplied in the prompt, and was required to be reproduced in the model output, allowing the evaluator to compare the returned identifier with the expected identifier stored externally for that generation attempt. Each relation used the fields \texttt{source}, \texttt{target}, and \texttt{relation}, with its label restricted to either \texttt{CONTRIBUTES\_TO} or \texttt{RESULTS\_IN}.

The following schematic illustrates the required structure; empty lists indicate categories without extracted items, while the relation object illustrates the required fields when a relation is returned:

\begin{verbatim}
{
  "record_id": "...",
  "event": [],
  "phase_of_flight": [],
  "hazard": [],
  "human_factor": [],
  "technical_factor": [],
  "environmental_factor": [],
  "contributing_factor": [],
  "consequence": [],
  "evidence": {
    "event": [],
    "phase_of_flight": [],
    "hazard": [],
    "human_factor": [],
    "technical_factor": [],
    "environmental_factor": [],
    "contributing_factor": [],
    "consequence": []
  },
  "relations": [
    {
      "source": "...",
      "target": "...",
      "relation": "CONTRIBUTES_TO"
    }
  ]
}
\end{verbatim}

For example, an evidence list could take the form \texttt{"phase\_of\_flight": ["On final approach"]}, with each string checked separately for verbatim occurrence in the corresponding narrative. This source-presence check did not independently establish whether the quotation supported the associated concept or category assignment. When no supported relations were extracted, the required value of \texttt{relations} was an empty list.

The schema remained constant across prompting conditions, and the instructions required extraction from the supplied narrative with evidence copied verbatim from that source. Concepts and evidence were represented separately so that semantic correctness and verbatim traceability could be evaluated independently.

\subsection{Language Model and Inference Configuration}

The main experiment used \texttt{HuggingFaceTB/SmolLM2-1.7B-Instruct}, an instruction-tuned model with approximately 1.7 billion parameters \cite{SmolLM2}. The same checkpoint was used under every prompting condition without task-specific fine-tuning.

Inference used an NVIDIA T4 graphics processing unit (GPU), 4-bit quantization with the NormalFloat4 (NF4) data type, 16-bit floating-point (FP16) computation, and double quantization \cite{Dettmers2023}. The model was placed in evaluation mode, and prompts were rendered using its chat template. Sampling was disabled through \texttt{do\_sample=False}, and generation was limited to 1,400 new tokens. These settings were held constant across conditions.

The experiment comprised 1,200 generation attempts: 300 narratives under each of four prompting conditions. Zero-shot inference was processed sequentially, whereas demonstration-based inference used a batch size of eight for computational efficiency. Batching changed execution organization rather than the intended prompt content, demonstration-selection rules, or scoring procedures.

\subsection{Prompting Conditions}

\subsubsection{Zero-Shot Prompting}

The zero-shot prompt contained extraction instructions, category and relation definitions, output requirements, and the target narrative, without an annotated demonstration. This condition evaluated extraction from task instructions alone.

\subsubsection{One-Shot Prompting}

One-shot prompting added the fixed annotated example \texttt{DEV\_01}. The same example was used for every benchmark narrative, illustrating concepts, evidence, relations, and output structure.

\subsubsection{Fixed Few-Shot Prompting}

Fixed few-shot prompting supplied four development examples: \texttt{DEV\_01}, \texttt{DEV\_05}, \texttt{DEV\_11}, and \texttt{DEV\_12}. Their selection remained unchanged across all 300 narratives. Four demonstrations were used to permit comparison with retrieval-grounded prompting at the same demonstration count.

\subsubsection{Retrieval-Grounded Prompting}

Retrieval-grounded prompting selected four demonstrations for each target narrative from the frozen pool of 12 examples. Narratives were encoded using \texttt{sentence-transformers/all-MiniLM-L6-v2}, producing 384-dimensional embeddings. Embeddings were normalized before cosine similarity was calculated.

The four development examples with the highest similarity to the target narrative were inserted into its prompt. Retrieval used narrative text only; neither reference annotations, \texttt{probableCause}, nor structured occurrence metadata were available to the selection mechanism. The candidate pool, embedding model, similarity procedure, and value of $k=4$ were fixed before inference.

Retrieval supplied annotated demonstrations rather than external factual evidence about the target occurrence. The comparison with fixed few-shot prompting therefore evaluated demonstration selection while holding the number of examples constant.

\subsection{Output Preservation and Structural Evaluation}

Raw generations were preserved before parsing or review. Structural evaluation assessed compliance with the requested machine-readable representation separately from semantic correctness.

The parser removed surrounding Markdown code fences when present and attempted to decode a JSON object. If this failed, it attempted to decode the substring between the first opening brace and the last closing brace. JSON parseability therefore indicates recovery of an object through this procedure, rather than strict validity of the entire raw response. Parsing did not modify the stored generation or repair extracted concepts and relations.

All 300 benchmark narratives were included in the evaluation of each condition, even when generation failed, the response could not be parsed, or the output reached the generation limit. When no JSON object could be recovered, the response contributed no concept predictions, and its reference concepts were counted as missed. Within recoverable objects, nonempty string predictions from available category lists were retained even if other structural requirements failed. The same rules applied to outputs reaching the generation limit, whose status was recorded separately.

Structural checks covered JSON recovery, record-ID correspondence, the required category lists, evidence-object and evidence-list structure, relation-list and relation-object structure, permitted relation labels, and endpoint validity. Complete structural compliance required all applicable checks to pass.

Unsupported labels were not mapped to the nearest permitted label, and malformed structures were not silently repaired. A parseable response could therefore fail complete structural compliance.

\subsection{Evidence-Grounding Evaluation}

Verbatim evidence grounding was assessed separately from concept accuracy. Each generated evidence span was tested for exact occurrence within its corresponding source narrative.

For $N_{\mathrm{evidence}}$ generated spans and $N_{\mathrm{exact}}$ verbatim matches, the grounding rate was

\begin{equation}
\mathrm{Grounding\ Rate}
=
\frac{N_{\mathrm{exact}}}
     {N_{\mathrm{evidence}}}
\times 100.
\end{equation}

The rate was reported as undefined when no usable evidence spans were generated. Spans failing the substring test were classified as non-verbatim for this measure; this designation does not independently establish whether their semantic content was faithful.

Verbatim occurrence establishes source traceability, not whether a span supports the associated concept, category, or relation. Concept correctness was evaluated separately, and no independent concept--evidence relevance assessment was included. The distinction between locating supporting material and assessing its relevance is consistent with broader evaluations of attribution and explanatory relevance \cite{Gao2023ALCE,Ieracitano2025TIxAI}; the present measure operationalizes only source-text presence.

\subsection{Concept-Level Semantic Evaluation}

Concept predictions were matched to reference concepts within the same narrative and category. One-to-one semantic matching allowed equivalent expressions to match despite lexical differences, while preventing a prediction or reference item from contributing to more than one accepted pair.

Candidate prediction--reference pairs were generated within each category. Embedding-based cosine similarity ranked candidates for review without applying a similarity threshold. Unambiguous normalized exact matches were locked before semantic review. Identical unresolved groups were reviewed once, and their decisions were mapped back to the corresponding prompting-condition entries.

The reconciled SmolLM2 checkpoint contained 2,225 unique concept-review tasks after deterministic resolution and removal of repeated groups. Of these, 132 were resolved through applicable historical adjudications and 2,093 through consolidated supplied decisions. The completed decisions resolved all 9,600 narrative--condition--category groups. These counts describe a shared, deduplicated workload rather than independently repeated judgments by each reviewer.

Reviewers were presented with the narrative, numbered predictions, numbered reference concepts, and any fixed matches. Matching considered meaning, category, polarity, intent, and relevant operational context. A broader prediction could contribute at most one match; its component phrases were not split into additional predictions during scoring. Missing defining information, contradictory intent, or a different event or operational phase could prevent a match even when expressions shared vocabulary.

Supplied reviewer sets and final adjudications were retained separately from exact matches and reused decisions. Contradictory pair fields or unresolved discussions were held until an explicit final decision was supplied. Minority judgments were preserved when the final adjudication rejected a pair by majority. Reuse did not create new independent reviewer votes.

Accepted pairs were counted as true positives (TP), unmatched predictions as false positives (FP), and unmatched reference concepts as false negatives (FN). The scoring code checked index validity and one-to-one correspondence before calculating metrics. No unresolved group was treated as an empty rejection.

\subsection{Relation-Level Evaluation}

Relation scoring required correspondence of the source concept, target concept, relation label, and direction. Concept identification alone was insufficient for a relation match.

Only \texttt{CONTRIBUTES\_TO} and \texttt{RESULTS\_IN} were permitted labels. Other labels and malformed relation objects remained model errors rather than being normalized or repaired. Such predictions were excluded from matching but retained in the FP count.

Relations were matched within each narrative and condition. Normalized exact matching required identical ordered endpoints and the same label. Remaining structurally valid predictions could match semantically equivalent endpoints only when their respective source and target roles, relation label, and direction were preserved.

The two relation labels were not interchangeable. Sharing a target consequence did not make different source events equivalent, and a prediction connecting an upstream condition directly to a downstream consequence was not credited for a reference link requiring a distinct intermediate event. Temporal association alone was insufficient to establish a causal or contributory match.

The reconciled SmolLM2 relation checkpoint contained 322 unique review tasks. Their completed decisions were mapped back to the 1,200 narrative--condition groups. The same recorded adjudication and correction procedures used for concepts were applied, with additional validation of relation labels and direction.

Each prediction and reference relation could contribute to at most one accepted pair. Accepted relations contributed to TP, unmatched generated relations, including malformed predictions, contributed to FP, and unmatched reference relations contributed to FN.

\subsection{Evaluation Metrics}

Precision, recall, and F1 were calculated as

\begin{equation}
\mathrm{Precision}
=
\frac{TP}{TP+FP},
\end{equation}

\begin{equation}
\mathrm{Recall}
=
\frac{TP}{TP+FN},
\end{equation}

and

\begin{equation}
F1
=
\frac{2TP}{2TP+FP+FN}.
\end{equation}

A metric with a zero denominator was assigned zero for semantic scoring.

Concept micro-averaged metrics aggregated counts across narratives and all eight categories before calculation. Macro-F1 was the unweighted mean of the eight category-level F1 scores, including categories with zero F1. Category-specific precision, recall, and F1 were also reported.

Relation metrics were calculated separately using all generated relations in the prediction denominator, including malformed generations retained as FP. Structural compliance and verbatim evidence grounding were reported as separate percentages and were not incorporated into semantic F1.

Uncertainty was quantified using 10,000 paired nonparametric bootstrap resamples with seed 42. For the main experiment, each replicate sampled 300 narrative identifiers with replacement and used the same sampled identifiers across all prompting conditions. All categories and relations associated with a sampled narrative were retained together.

Semantic metrics were recomputed from aggregated counts within each replicate. Macro-F1 was recomputed as the mean of the eight resampled category-level F1 scores, with zero assigned when a category denominator was zero. For each bootstrap replicate, we recalculated the grounding rate by dividing the number of verbatim spans by the total number of generated evidence spans. We calculated complete structural compliance as the proportion of sampled generation attempts that yielded an output passing all applicable checks. 

Percentile 95\% confidence intervals used the 2.5th and 97.5th percentiles of each bootstrap distribution. Differences between conditions were calculated within paired replicates and expressed in percentage points. Intervals were calculated separately without adjustment for multiple comparisons.

All semantic metrics and their intervals use the completed reconciled reference and finalized matching decisions. The intervals describe narrative-sampling variability conditional on those decisions; they do not quantify annotation uncertainty, adjudication uncertainty, or uncertainty arising from model selection.

\subsection{Narrative-Length Analysis}

Concept extraction was additionally evaluated within the predefined short, medium, and long strata. The benchmark contained 76 short narratives with at most 92 words, 150 medium narratives with 93--195 words, and 74 long narratives with more than 195 words.

For each condition and stratum, concept TP, FP, and FN were aggregated across all eight categories, and micro-precision, micro-recall, and micro-F1 were calculated. The reconciled reference contained 759 concepts in short narratives, 1,902 in medium narratives, and 1,102 in long narratives.

This analysis was descriptive. Narrative length was not independently manipulated, and its association with extraction performance could also reflect differences in concept density, event complexity, or other report characteristics.

\subsection{Reproducibility and Evaluation Controls}

The schema, development examples, prompting conditions, and benchmark membership were fixed before inference. The pilot cases and development examples were excluded from the 300-case benchmark, which was sampled using seed 42.

Fixed demonstrations remained unchanged across target narratives. Retrieval-selected demonstrations came exclusively from the frozen development pool, using narrative similarity without access to benchmark annotations or withheld fields. Model outputs were retained without post-hoc content correction.

Reference reconciliation changed the scoring annotations rather than the benchmark narratives or model generations. The completed reconciled reference was fixed for final scoring, and all conditions were evaluated against it. Prior matching decisions were reused only where applicable, with reuse provenance distinguished from new decisions. Superseded submissions and explicit corrections were retained to support traceability.

The scoring pipeline verified valid indices, one-to-one matching, relation-label correspondence, consistent reference counts, and the absence of unresolved groups before calculating final results. Structural compliance and verbatim grounding remained separate from semantic scoring.

\subsection{Exploratory Comparison with a Larger Language Model}
\label{sec:qwen-supplementary}

The supplementary experiment compared SmolLM2-1.7B-Instruct with \texttt{Qwen/Qwen3-8B} in non-thinking mode \cite{Yang2025Qwen3}. It used the first 35 benchmark narratives processed during the additional inference run, \texttt{BENCH\_001}--\texttt{BENCH\_035}, comprising eight short, 17 medium, and ten long narratives.

The subset was selected by processing order under computational constraints rather than by random or stratified sampling. Qwen3-8B generated 140 outputs across the 35 narratives and four conditions. SmolLM2 was evaluated using its corresponding existing outputs, ensuring identical narrative and condition coverage without rerunning the main model.

Qwen3-8B used the same prompt contents and demonstration selections, rendered through its native chat template with thinking disabled. Inference used NF4 quantization, double quantization, FP16 computation, greedy decoding, and a maximum of 1,400 new tokens on an NVIDIA Tesla T4 GPU.

Both models were scored against the same reconciled subset reference containing 432 concepts and 77 relations. Structural compliance and verbatim grounding were evaluated from their saved outputs and source narratives using the procedures applied in the main experiment.

Applicable prior Qwen decisions, normalized exact matches, and reconciled SmolLM2 matches were reused with recorded provenance. Following reuse and removal of repeated groups, 374 additional unique tasks required review: 295 concept tasks and 79 relation tasks. All were resolved, including explicit correction of contradictory pair fields and rejection of source, target, direction, or relation-label mismatches. Reused decisions were not represented as new independent reviewer judgments.

The completed Qwen evaluation comprised 1,120 concept groups and 140 relation groups. Every group was resolved before scoring, and both models used identical reference counts for each condition. SmolLM2 subset scores were recomputed rather than taken from its full 300-case aggregate.

Paired percentile intervals used 10,000 resamples of the same 35 narrative identifiers across both models and all conditions, with seed 42. Within each replicate, metrics were recomputed from aggregated counts, and Qwen3-8B minus SmolLM2 differences were calculated. Intervals were unadjusted for multiple comparisons and conditional on the finalized matching decisions.

The comparison is exploratory. Its small subset and selection by processing order restrict generalization to the full benchmark. Differences in architecture, training, and parameter count also prevent attribution of observed performance differences specifically to model size.

\section{Results}
\label{sec:results}

All semantic extraction results in this section were recomputed against the completed reconciled reference containing 3,763 concepts and 467 directed relations across 300 narratives.
The supplementary model comparison uses the corresponding 35-narrative reference containing 432 concepts and 77 directed
relations. Semantic scores and paired bootstrap intervals reflect the finalized one-to-one matching decisions.
Generation outcomes, structural compliance, and verbatim evidence-grounding measurements remain unchanged because
they depend on the saved model outputs and source narratives rather than the reference annotation version.

\subsection{Generation Outcomes}

The main SmolLM2 experiment comprised 1,200 generation attempts, corresponding to 300 benchmark narratives under each of four prompting conditions. Of these, 1,192 returned generated responses, while eight generation failures occurred, all under retrieval-grounded prompting. Returning a response did not imply successful JSON parsing, structural compliance, or termination before the generation limit.

The maximum generation limit of 1,400 new tokens was reached in 302 of the 1,200 generation attempts (25.2\%). This behavior differed substantially across prompting conditions. In the zero-shot condition, 213 of 300 generations (71.0\%) reached the generation ceiling, compared with 38 of 300 (12.7\%) for one-shot prompting, 27 of 300 (9.0\%) for few-shot prompting, and 24 of 300 (8.0\%) for retrieval-grounded prompting. Although the zero-shot prompts were shorter and contained no demonstrations, the model reached the generation limit more often under this condition than under any of the other three. 

\subsection{Reference-Annotation Agreement}
\label{sec:reference-agreement-results}

Following discussion and reconciliation, all three reviewers confirmed the final annotations for all 300 narratives. These confirmations document approval of the consensus reference and were recorded separately from the individual submissions used to assess pre-consensus agreement.

Three-reviewer agreement diagnostics were calculated for 232 narratives with usable, unambiguous submissions and matching case identifiers. The remaining 68 narratives were excluded because 57 retained Reviewer 3 submissions contained case-ID mismatches, ten contained multiple submitted versions, and one contained invalid annotations. Final consensus decisions were not substituted for missing or unusable independent judgments.

Table~\ref{tab:reference-agreement} reports agreement on whether each category contained at least one annotation. Three-reviewer unanimity includes both unanimous presence and unanimous absence. Fleiss' $\kappa$ was calculated separately for each category, with percentile 95\% confidence intervals obtained from 10,000 narrative-level bootstrap resamples using random seed 42. These measures assess category presence rather than semantic agreement on individual concepts, evidence spans, or complete directed relations.

\begin{table}[H]
\caption{Pre-consensus category-presence agreement across 232 eligible narratives. Confidence intervals are narrative-level bootstrap intervals and are not adjusted for multiple comparisons.}
\label{tab:reference-agreement}
\centering
\small
\setlength{\tabcolsep}{5pt}
\begin{tabular}{lrrl}
\toprule
\textbf{Category} & \textbf{Unanimity (\%)} & \textbf{Fleiss' $\kappa$} & \textbf{95\% CI} \\
\midrule
Event                & 94.40 & 0.052 & $-0.025$ to $0.188$ \\
Phase of flight      & 93.97 & 0.279 & $-0.027$ to $0.561$ \\
Hazard               & 70.26 & 0.494 & $0.382$ to $0.593$ \\
Human factor         & 83.62 & 0.772 & $0.704$ to $0.834$ \\
Technical factor     & 91.38 & 0.858 & $0.792$ to $0.913$ \\
Environmental factor & 59.48 & 0.456 & $0.366$ to $0.539$ \\
Contributing factor  & 94.83 & 0.059 & $-0.024$ to $0.198$ \\
Consequence          & 93.10 & 0.689 & $0.529$ to $0.816$ \\
Any relation         & 65.09 & 0.519 & $0.431$ to $0.601$ \\
\bottomrule
\end{tabular}
\end{table}

Technical-factor presence produced the highest observed $\kappa$ of 0.858, followed by human-factor presence at 0.772 and consequence presence at 0.689. Hazard and environmental-factor presence yielded $\kappa$ values of 0.494 and 0.456, respectively, while agreement on whether any relation was annotated yielded a $\kappa$ of 0.519.

High unanimity did not always correspond to high chance-corrected agreement because category-presence frequencies were strongly imbalanced. Event presence had a unanimity rate of 94.40\% but a $\kappa$ of 0.052. Contributing-factor presence similarly had a unanimity rate of 94.83\% but a $\kappa$ of 0.059; all three reviewers marked contributing factors absent in 220 narratives, and no narrative had unanimous contributing-factor presence. The coefficients are therefore interpreted alongside unanimity and according to their limited category-presence scope. These diagnostics concern reference category presence only. The completed semantic-matching records do not establish uniformly independent pre-adjudication judgments across the expanded evaluation; consequently, no independent
semantic-matching agreement coefficient is reported for the reconciled results.

\subsection{Concept-Level Extraction Performance}

Across the four prompting conditions, SmolLM2 generated 5,183 concept predictions, which were evaluated against the same reconciled reference containing 3,763 concepts per condition, yielding 15,052 reference instances across the four evaluations. Finalized one-to-one matching identified 2,590 true-positive concept predictions, 2,593 false positives, and 12,462 false negatives.

Table~\ref{tab:concept_overall} summarizes concept-level performance across the four prompting conditions. Zero-shot prompting achieved a micro-F1 of 3.72\%, whereas introducing one annotated demonstration increased micro-F1 to 30.75\%, with further increases to 31.52\% under fixed few-shot prompting and 32.27\% under retrieval-grounded prompting. Retrieval-grounded prompting achieved the highest observed precision, recall, and micro-F1, while its macro-F1 of 20.76\% was close to the 20.73\% achieved under fixed few-shot prompting.

\begin{table}[htbp]
\caption{Concept-level extraction performance against the reconciled reference containing 3,763 concepts per condition. Precision, recall, micro-F1, and macro-F1 are percentages.}
\label{tab:concept_overall}
\centering
\small
\setlength{\tabcolsep}{3pt}
\begin{tabular}{lrrrrrrrr}
\toprule
\textbf{Condition} & \textbf{Pred.} & \textbf{TP} & \textbf{FP} & \textbf{FN} & \textbf{P} & \textbf{R} & \textbf{Micro-F1} & \textbf{Macro-F1} \\
\midrule
Zero-shot          & 540  & 80  & 460 & 3683 & 14.81 & 2.13  & 3.72           & 3.72 \\
One-shot           & 1603 & 825 & 778 & 2938 & 51.47 & 21.92 & 30.75          & 17.87 \\
Fixed few-shot     & 1504 & 830 & 674 & 2933 & 55.19 & 22.06 & 31.52          & 20.73 \\
Retrieval-grounded & 1536 & 855 & 681 & 2908 & \textbf{55.66} & \textbf{22.72} & \textbf{32.27} & \textbf{20.76} \\
\bottomrule
\end{tabular}
\end{table}

Paired narrative-level bootstrap analysis yielded concept micro-F1 estimates of 3.72\% (95\% CI: 2.51--5.03) for zero-shot prompting, 30.75\% (28.60--32.91) for one-shot prompting, 31.52\% (29.72--33.34) for fixed few-shot prompting, and 32.27\% (30.42--34.11) for retrieval-grounded prompting. The paired retrieval-grounded-minus-fixed-few-shot difference was 0.75 percentage points (95\% CI: $-0.92$ to $2.40$), indicating that the higher retrieval-grounded point estimate did not establish a consistent micro-F1 advantage over the fixed demonstration set.

The corresponding macro-F1 estimates were 3.72\% (95\% CI: 2.52--5.02), 17.87\% (16.70--19.05), 20.73\% (19.47--22.04), and 20.76\% (19.41--22.14) for zero-shot, one-shot, fixed few-shot, and retrieval-grounded prompting, respectively. Calculated from unrounded scores, the paired macro-F1 difference between retrieval-grounded and fixed few-shot prompting was 0.02 percentage points (95\% CI: $-1.36$ to $1.39$), providing insufficient evidence of an advantage for either four-demonstration strategy on this measure.

\begin{figure}[htbp]
\centering
\begin{tikzpicture}
\begin{axis}[
    makeaxis,
    ybar,
    bar width=9pt,
    ymin=0,
    ymax=65,
    ylabel={Score (\%)},
    symbolic x coords={Zero,One,Fixed,Retrieved},
    xtick=data,
    xticklabels={
        Zero-shot,
        One-shot,
        Fixed few-shot,
        Retrieval-grounded
    },
    x tick label style={
        font=\footnotesize,
        text width=2.3cm,
        align=center
    },
    legend style={
        at={(0.5,1.03)},
        anchor=south,
        legend columns=3,
        draw=none,
        fill=none,
        font=\footnotesize
    }
]

\addplot[
    fill=makeblue,
    draw=makeblue
] coordinates {
    (Zero,14.81)
    (One,51.47)
    (Fixed,55.19)
    (Retrieved,55.66)
};

\addplot[
    fill=makeorange,
    draw=makeorange
] coordinates {
    (Zero,2.13)
    (One,21.92)
    (Fixed,22.06)
    (Retrieved,22.72)
};

\addplot[
    fill=makegreen,
    draw=makegreen
] coordinates {
    (Zero,3.72)
    (One,30.75)
    (Fixed,31.52)
    (Retrieved,32.27)
};

\legend{Precision,Recall,Micro-F1}

\end{axis}
\end{tikzpicture}
\caption{Concept-level precision, recall, and micro-F1 across
the four prompting conditions, evaluated against the reconciled
reference containing 3,763 concepts across 300 narratives.
All values are percentages.}
\label{fig:concept_performance}
\end{figure}
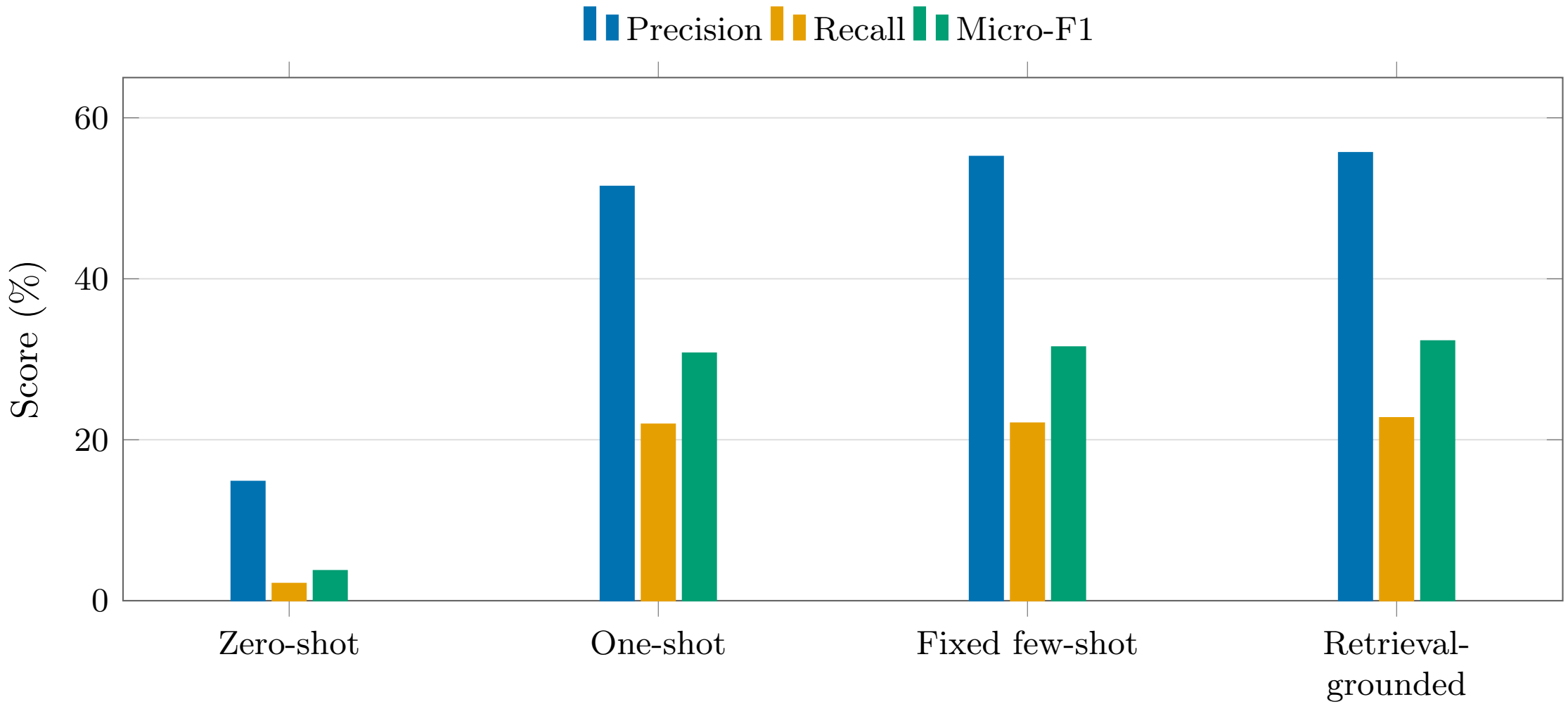

Figure~\ref{fig:concept_performance} shows that concept micro-F1 increased from 3.72\% under zero-shot prompting to 30.75\% under one-shot prompting. Differences among the three demonstration-based conditions were smaller. Recall remained below 23\% under all four conditions.

\subsection{Category-Level Concept Extraction}

Table~\ref{tab:retrieval-coverage} summarizes the selection frequency and concept-category coverage of the 12 development examples across the 300 retrieval-grounded prompts. The four retrieved examples provided an average of 2.03 environmental-factor concepts per prompt, compared with one in the fixed four-example set, while the average number of human-factor concepts was 2.15 under retrieval and 4.00 under fixed few-shot prompting. At least one example containing an environmental-factor concept appeared in 296 of the 300 retrieval-grounded prompts. Selection frequencies were uneven, ranging from 26 appearances for DEV\_10 to 229 for DEV\_07 across the 1,200 demonstration slots.

\begin{table}[htbp]
\caption{Development-example selection frequencies and annotated concept counts by category. The fixed few-shot condition used DEV\_01, DEV\_05, DEV\_11, and DEV\_12 in every prompt.}
\label{tab:retrieval-coverage}
\centering
\small
\begin{tabular}{lrrrrrrrrr}
\toprule
Example & Retrieved & E & P & H & HF & TF & EF & CF & C \\
\midrule
DEV\_01 & 118 & 1 & 1 & 1 & 0 & 0 & 0 & 0 & 4 \\
DEV\_02 &  86 & 1 & 1 & 1 & 1 & 0 & 0 & 0 & 3 \\
DEV\_03 & 218 & 1 & 2 & 1 & 0 & 0 & 0 & 0 & 2 \\
DEV\_04 &  44 & 1 & 1 & 1 & 0 & 0 & 1 & 0 & 2 \\
DEV\_05 &  78 & 1 & 1 & 1 & 1 & 0 & 1 & 0 & 3 \\
DEV\_06 & 149 & 1 & 1 & 1 & 0 & 0 & 1 & 0 & 3 \\
DEV\_07 & 229 & 1 & 2 & 1 & 1 & 0 & 1 & 0 & 5 \\
DEV\_08 &  92 & 1 & 1 & 1 & 1 & 0 & 0 & 0 & 2 \\
DEV\_09 &  55 & 1 & 2 & 2 & 0 & 0 & 2 & 0 & 3 \\
DEV\_10 &  26 & 1 & 1 & 3 & 0 & 2 & 0 & 1 & 4 \\
DEV\_11 &  49 & 1 & 2 & 1 & 1 & 0 & 0 & 0 & 2 \\
DEV\_12 &  56 & 1 & 2 & 1 & 2 & 0 & 0 & 0 & 3 \\
\bottomrule
\end{tabular}

\medskip
\begin{minipage}{\linewidth}
\footnotesize
Retrieved denotes the number of benchmark prompts selecting each example, summed over all four retrieval ranks. E denotes event, P phase of flight, H hazard, HF human factor, TF technical factor, EF environmental factor, CF contributing factor, and C consequence. Category columns report annotated concept instances within each development example.
\end{minipage}
\end{table}

Performance varied substantially across the eight concept categories, as shown in Table~\ref{tab:category_results}, with different prompting conditions achieving the highest observed F1 for different categories. The reconciled reference contained 1,285 events, 613 phases of flight, 536 hazards, 212 human factors, 154 technical factors, 394 environmental factors, nine contributing factors, and 560 consequences.

\begin{table}[htbp]
\caption{Category-level precision, recall, and F1 against the reconciled reference across the four prompting conditions. All values are percentages.}
\label{tab:category_results}
\centering
\small
\setlength{\tabcolsep}{6pt}
\begin{tabular}{llrrr}
\toprule
\textbf{Condition} & \textbf{Category} & \textbf{Precision} & \textbf{Recall} & \textbf{F1} \\
\midrule
Zero-shot & Event & 0.00 & 0.00 & 0.00 \\
          & Phase of flight & 29.41 & 4.89 & 8.39 \\
          & Hazard & 8.14 & 1.31 & 2.25 \\
          & Human factor & 7.69 & 2.36 & 3.61 \\
          & Technical factor & 4.35 & 1.95 & 2.69 \\
          & Environmental factor & 23.81 & 3.81 & 6.56 \\
          & Contributing factor & 0.00 & 0.00 & 0.00 \\
          & Consequence & 25.00 & 3.57 & 6.25 \\
\midrule
One-shot & Event & 74.39 & 21.48 & 33.33 \\
         & Phase of flight & 70.47 & 34.26 & 46.10 \\
         & Hazard & 32.38 & 16.98 & 22.28 \\
         & Human factor & 0.00 & 0.00 & 0.00 \\
         & Technical factor & 0.00 & 0.00 & 0.00 \\
         & Environmental factor & 100.00 & 0.25 & 0.51 \\
         & Contributing factor & 0.00 & 0.00 & 0.00 \\
         & Consequence & 37.88 & 44.11 & 40.76 \\
\midrule
Fixed few-shot & Event & 86.48 & 18.91 & 31.03 \\
               & Phase of flight & 77.27 & 38.83 & 51.68 \\
               & Hazard & 30.08 & 13.81 & 18.93 \\
               & Human factor & 17.02 & 15.09 & 16.00 \\
               & Technical factor & 100.00 & 0.65 & 1.29 \\
               & Environmental factor & 50.00 & 0.25 & 0.51 \\
               & Contributing factor & 0.00 & 0.00 & 0.00 \\
               & Consequence & 50.42 & 43.04 & 46.44 \\
\midrule
Retrieval-grounded & Event & 86.73 & 19.84 & 32.30 \\
                   & Phase of flight & 78.79 & 38.17 & 51.43 \\
                   & Hazard & 30.55 & 15.67 & 20.72 \\
                   & Human factor & 14.58 & 3.30 & 5.38 \\
                   & Technical factor & 0.00 & 0.00 & 0.00 \\
                   & Environmental factor & 57.14 & 7.11 & 12.64 \\
                   & Contributing factor & 0.00 & 0.00 & 0.00 \\
                   & Consequence & 43.11 & 44.11 & 43.60 \\
\bottomrule
\end{tabular}
\end{table}

Phase of flight achieved the highest category-level F1 under both fixed few-shot and retrieval-grounded prompting, reaching 51.68\% and 51.43\%, respectively, while consequence extraction achieved its highest F1 of 46.44\% under fixed few-shot prompting. One-shot prompting achieved the highest observed event F1 of 33.33\% and hazard F1 of 22.28\%, demonstrating that increasing the number of demonstrations did not consistently improve every category.

Retrieval-grounded prompting achieved the highest environmental-factor F1 of 12.64\%, compared with 0.51\% under fixed few-shot prompting, whereas human-factor F1 was higher under fixed few-shot prompting at 16.00\% than under retrieval-grounded prompting at 5.38\%. These descriptive differences accompanied the differences in demonstration coverage reported in Table~\ref{tab:retrieval-coverage}, although category coverage was not independently manipulated and cannot be identified as their cause.

High precision in sparsely predicted categories did not imply broad reference coverage: the 100.00\% technical-factor precision under fixed few-shot prompting arose from one correct prediction against 154 reference instances, yielding recall of 0.65\% and F1 of 1.29\%, while the 100.00\% environmental-factor precision under one-shot prompting arose from one correct prediction against 394 reference instances, yielding recall of 0.25\% and F1 of 0.51\%. No contributing-factor instance was correctly recovered under any prompting condition, although the reference contained only nine instances of this category.

Figure~\ref{fig:category_heatmap} summarizes category-level F1 across the four prompting conditions, illustrating the variation that is obscured when performance is represented only through aggregate micro- or macro-F1.

\begin{figure}[htbp]
\centering
\begin{tikzpicture}[x=1.55cm,y=0.58cm]

\foreach \x/\heading in {
    0/Zero-shot,
    1/One-shot,
    2/Fixed few-shot,
    3/Retrieval-grounded
}{
    \node[
        font=\scriptsize,
        align=center,
        text width=1.5cm,
        anchor=south
    ] at (\x,0.55) {\heading};
}

\foreach \y/\category in {
    0/Event,
    1/Phase of flight,
    2/Hazard,
    3/Human factor,
    4/Technical factor,
    5/Environmental factor,
    6/Contributing factor,
    7/Consequence
}{
    \node[
        anchor=east,
        font=\scriptsize
    ] at (-0.55,-\y) {\category};
}

\foreach \x/\y/\value in {
    0/0/0.00, 1/0/33.33, 2/0/31.03, 3/0/32.30,
    0/1/8.39, 1/1/46.10, 2/1/51.68, 3/1/51.43,
    0/2/2.25, 1/2/22.28, 2/2/18.93, 3/2/20.72,
    0/3/3.61, 1/3/0.00, 2/3/16.00, 3/3/5.38,
    0/4/2.69, 1/4/0.00, 2/4/1.29, 3/4/0.00,
    0/5/6.56, 1/5/0.51, 2/5/0.51, 3/5/12.64,
    0/6/0.00, 1/6/0.00, 2/6/0.00, 3/6/0.00,
    0/7/6.25, 1/7/40.76, 2/7/46.44, 3/7/43.60
}{
    \pgfmathtruncatemacro{\shade}{100*\value/60}
    \filldraw[
        fill=blue!\shade!white,
        draw=black!15,
        line width=0.3pt
    ]
        (\x-0.5,-\y-0.5)
        rectangle
        (\x+0.5,-\y+0.5);

    \ifdim\value pt>35pt
        \node[font=\scriptsize,text=white]
            at (\x,-\y) {\value};
    \else
        \node[font=\scriptsize,text=black]
            at (\x,-\y) {\value};
    \fi
}

\node[
    font=\scriptsize,
    anchor=north,
    align=center
] at (1.5,-7.8)
    {F1 (\%); darker shading indicates higher scores.};

\end{tikzpicture}
\caption{Category-level concept F1 across the four prompting
conditions, evaluated against the reconciled reference.
Cell values are percentages.}
\label{fig:category_heatmap}
\end{figure}

\subsection{Concept Extraction Performance by Narrative Length }

Concept-extraction performance decreased across the short, medium, and long narrative strata under all four prompting conditions, as summarized in Table~\ref{tab:length-results}. The strata contained 76 short narratives of at most 92 words, 150 medium narratives of 93--195 words, and 74 long narratives exceeding 195 words, with 759, 1,902, and 1,102 reconciled reference concepts, respectively.

\begin{table}[htbp]
\caption{Concept micro-F1 across narrative-length strata, evaluated against the reconciled reference. All performance values are percentages.}
\label{tab:length-results}
\centering
\small
\setlength{\tabcolsep}{8pt}
\begin{tabular}{lrrr}
\toprule
\textbf{Condition} & \textbf{Short} & \textbf{Medium} & \textbf{Long} \\
\midrule
Zero-shot          & 10.34 & 2.41  & 0.52 \\
One-shot           & \textbf{43.86} & 29.39 & 22.67 \\
Fixed few-shot     & 42.70 & \textbf{31.97} & 21.90 \\
Retrieval-grounded & 43.59 & 31.49 & \textbf{24.69} \\
\midrule
Narratives         & 76  & 150  & 74 \\
Reference concepts & 759 & 1902 & 1102 \\
\bottomrule
\end{tabular}
\end{table}

Under zero-shot prompting, micro-F1 decreased from 10.34\% for short narratives to 2.41\% for medium narratives and 0.52\% for long narratives, while the corresponding scores under one-shot prompting were 43.86\%, 29.39\%, and 22.67\%. The same declining pattern occurred under fixed few-shot prompting, with scores of 42.70\%, 31.97\%, and 21.90\%, and under retrieval-grounded prompting, with scores of 43.59\%, 31.49\%, and 24.69\%.

The decline in F1 was accompanied by reduced recall, with retrieval-grounded recall decreasing from 32.94\% for short narratives to 22.50\% for medium narratives and 16.06\% for long narratives. Thus, even under the condition with the highest overall concept micro-F1, a progressively smaller proportion of reference concepts was recovered across increasing narrative-length strata.

The highest observed micro-F1 differed by stratum, with one-shot prompting performing best for short narratives, fixed few-shot prompting for medium narratives, and retrieval-grounded prompting for long narratives. These comparisons are descriptive and do not establish statistical superiority within individual strata, while the association between narrative length and performance cannot be separated from differences in concept density, event complexity, or other narrative characteristics through this analysis alone.

\begin{figure}[htbp]
\centering
\begin{tikzpicture}
\begin{axis}[
    makeaxis,
    ymin=0,
    ymax=50,
    ylabel={Concept micro-F1 (\%)},
    xlabel={Narrative-length group},
    symbolic x coords={Short,Medium,Long},
    xtick=data,
    xticklabels={
        Short,
        Medium,
        Long
    },
    legend style={
        at={(0.5,1.03)},
        anchor=south,
        legend columns=2,
        draw=none,
        fill=none,
        font=\footnotesize
    }
]

\addplot[
    color=black!60,
    mark=*,
    thick
] coordinates {
    (Short,10.34)
    (Medium,2.41)
    (Long,0.52)
};

\addplot[
    color=makeorange,
    mark=square*,
    thick
] coordinates {
    (Short,43.86)
    (Medium,29.39)
    (Long,22.67)
};

\addplot[
    color=makeblue,
    mark=triangle*,
    thick
] coordinates {
    (Short,42.70)
    (Medium,31.97)
    (Long,21.90)
};

\addplot[
    color=makegreen,
    mark=diamond*,
    thick
] coordinates {
    (Short,43.59)
    (Medium,31.49)
    (Long,24.69)
};

\legend{
    Zero-shot,
    One-shot,
    Fixed few-shot,
    Retrieval-grounded
}

\end{axis}
\end{tikzpicture}
\caption{Concept micro-F1 across short ($\leq92$ words;
$n=76$), medium (93--195 words; $n=150$), and long
($>195$ words; $n=74$) narratives, evaluated against the
reconciled reference. Lines connect descriptive stratum-level
estimates and do not represent a longitudinal trend.}
\label{fig:concept-length}
\end{figure}
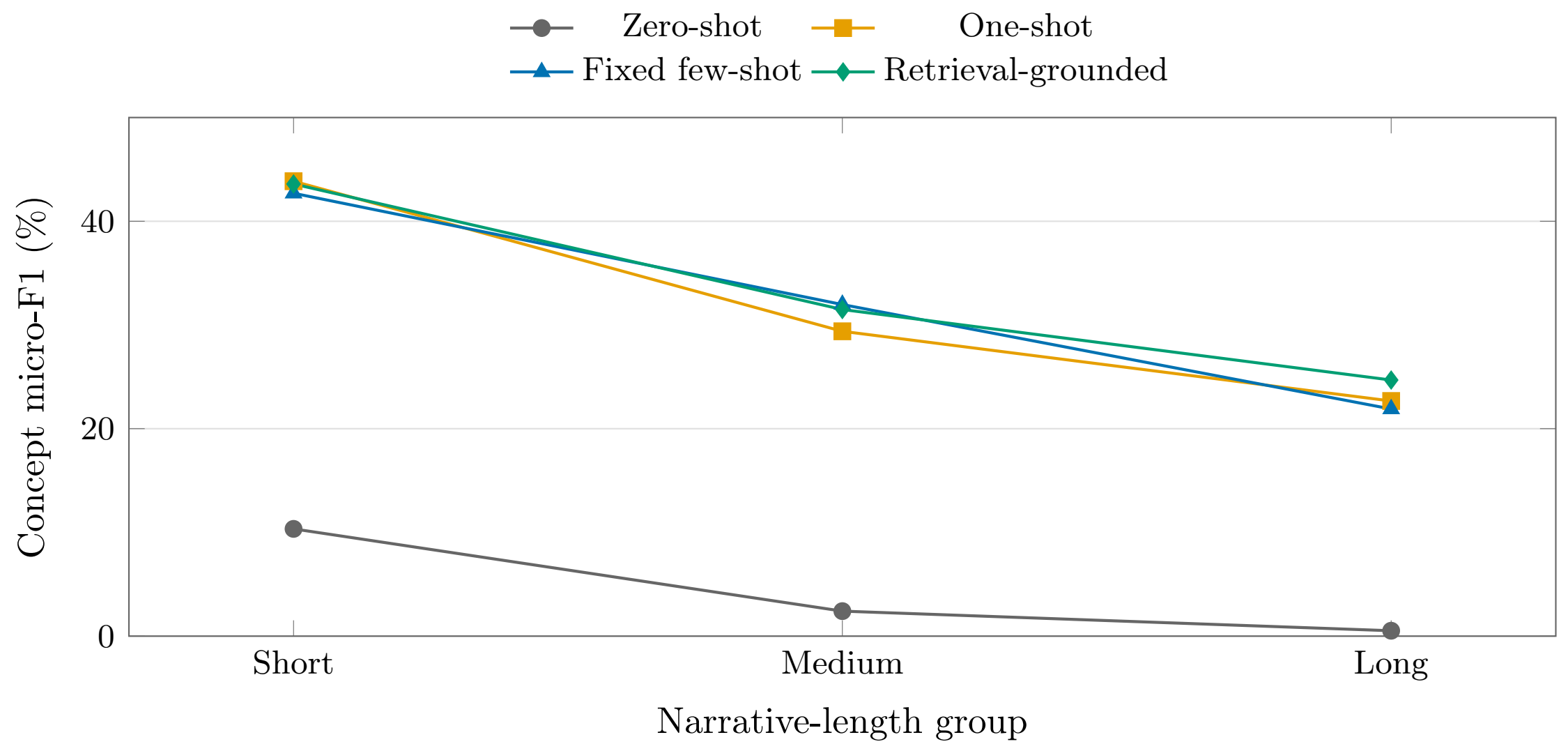

Figure~\ref{fig:concept-length} illustrates these stratum-level patterns across the four prompting conditions.

\subsection{Structural Compliance and Evidence Grounding}

Structural compliance and verbatim evidence grounding differed across prompting conditions, as summarized in Table~\ref{tab:structural-grounding}, with JSON parseability consistently exceeding complete compliance because a parseable response could still violate the required extraction schema. These measurements remain unchanged following reference reconciliation because they were calculated from the saved model outputs and source narratives.

\begin{table}[htbp]
\caption{Structural compliance and verbatim evidence grounding across the four prompting conditions. JSON parsing and complete compliance are reported relative to 300 generation attempts per condition, while grounding is the percentage of generated evidence spans occurring verbatim in the corresponding source narrative.}
\label{tab:structural-grounding}
\centering
\small
\setlength{\tabcolsep}{3pt}
\begin{tabular}{lrrrrr}
\toprule
\textbf{Condition}
& \shortstack{\textbf{JSON}\\\textbf{parsed}}
& \shortstack{\textbf{Complete}\\\textbf{compliance}}
& \shortstack{\textbf{Generated}\\\textbf{spans}}
& \shortstack{\textbf{Verbatim}\\\textbf{spans}}
& \shortstack{\textbf{Grounding}\\\textbf{(\%)}} \\
\midrule
Zero-shot          & 71/300  & 29/300 (9.7\%)   & 0    & 0   & -- \\
One-shot           & 256/300 & 160/300 (53.3\%) & 1555 & 806 & 51.83 \\
Fixed few-shot     & 272/300 & 128/300 (42.7\%) & 1516 & 819 & 54.02 \\
Retrieval-grounded & 268/300 & 100/300 (33.3\%) & 1519 & 903 & 59.45 \\
\bottomrule
\end{tabular}
\end{table}

Zero-shot prompting produced 71 parseable responses, but only 29 of the 300 attempts satisfied all structural requirements, while no usable evidence spans were available for calculating verbatim grounding. Its grounding rate is therefore undefined rather than zero, distinguishing the absence of evaluable spans from the generation of evidence that fails the substring check. Thus, passing all applicable structural checks did not guarantee that a response contained a usable evidence span. 

One-shot prompting achieved the highest complete structural compliance, with 160 compliant outputs out of 300 attempts, whereas fixed few-shot and retrieval-grounded prompting produced 128 and 100 compliant outputs, respectively. Fixed few-shot prompting achieved the highest JSON parseability at 272 of 300 attempts, demonstrating that successful parsing alone did not establish adherence to the complete schema.

Among the demonstration-based conditions, the proportion of generated evidence spans appearing verbatim in the source narrative increased from 51.83\% under one-shot prompting to 54.02\% under fixed few-shot prompting and 59.45\% under retrieval-grounded prompting. The corresponding numbers of non-verbatim spans were 749, 697, and 616, respectively; these spans failed the exact source-presence criterion, although that failure alone does not determine whether they were faithful paraphrases or unsupported statements.

Paired narrative-level bootstrap analysis yielded a retrieval-grounded verbatim-grounding rate of 59.45\% (95\% CI: 56.13--62.72), compared with 54.02\% (50.89--57.09) under fixed few-shot prompting, giving a paired difference of 5.42 percentage points (95\% CI: 1.89--8.85). Complete structural compliance was 33.33\% (95\% CI: 28.00--38.67) under retrieval-grounded prompting and 42.67\% (37.00--48.33) under fixed few-shot prompting, giving a paired difference of $-9.33$ percentage points (95\% CI: $-17.00$ to $-1.33$).

These individual, unadjusted intervals support an observed trade-off between more frequent verbatim evidence spans and fewer completely compliant outputs under retrieval-grounded prompting relative to fixed few-shot prompting. However, the grounding calculation establishes only that a quoted span occurs in the narrative and does not determine whether it supports the associated concept, justifies its category assignment, or establishes a proposed relation.

\begin{figure}[htbp]
\centering
\begin{tikzpicture}
\begin{axis}[
    makeaxis,
    ybar,
    bar width=18pt,
    ymin=0,
    ymax=75,
    ylabel={Rate (\%)},
    symbolic x coords={One,Fixed,Retrieved},
    xtick=data,
    xticklabels={
        One-shot,
        Fixed few-shot,
        Retrieval-grounded
    },
    x tick label style={
        font=\footnotesize,
        text width=2.5cm,
        align=center
    },
    enlarge x limits=0.25,
    nodes near coords,
    nodes near coords align={vertical},
    every node near coord/.append style={
        font=\scriptsize,
        /pgf/number format/fixed,
        /pgf/number format/precision=2,
        /pgf/number format/zerofill
    },
    legend style={
        at={(0.5,1.03)},
        anchor=south,
        legend columns=1,
        draw=none,
        fill=none,
        font=\footnotesize
    }
]

\addplot[
    fill=makepurple,
    draw=makepurple
] coordinates {
    (One,53.33)
    (Fixed,42.67)
    (Retrieved,33.33)
};

\addplot[
    fill=makegreen,
    draw=makegreen
] coordinates {
    (One,51.83)
    (Fixed,54.02)
    (Retrieved,59.45)
};

\legend{
    Complete structural compliance,
    Verbatim evidence grounding
}

\end{axis}
\end{tikzpicture}
\caption{Complete structural compliance and verbatim evidence
grounding across the demonstration-based prompting conditions.
Compliance is the percentage of generation attempts satisfying
the complete schema, whereas grounding is the percentage of
generated evidence spans occurring verbatim in the source
narrative. Zero-shot prompting is omitted because its grounding
rate is undefined in the absence of usable evidence spans.
Verbatim presence does not establish evidential relevance.}
\label{fig:structural_grounding}
\end{figure}
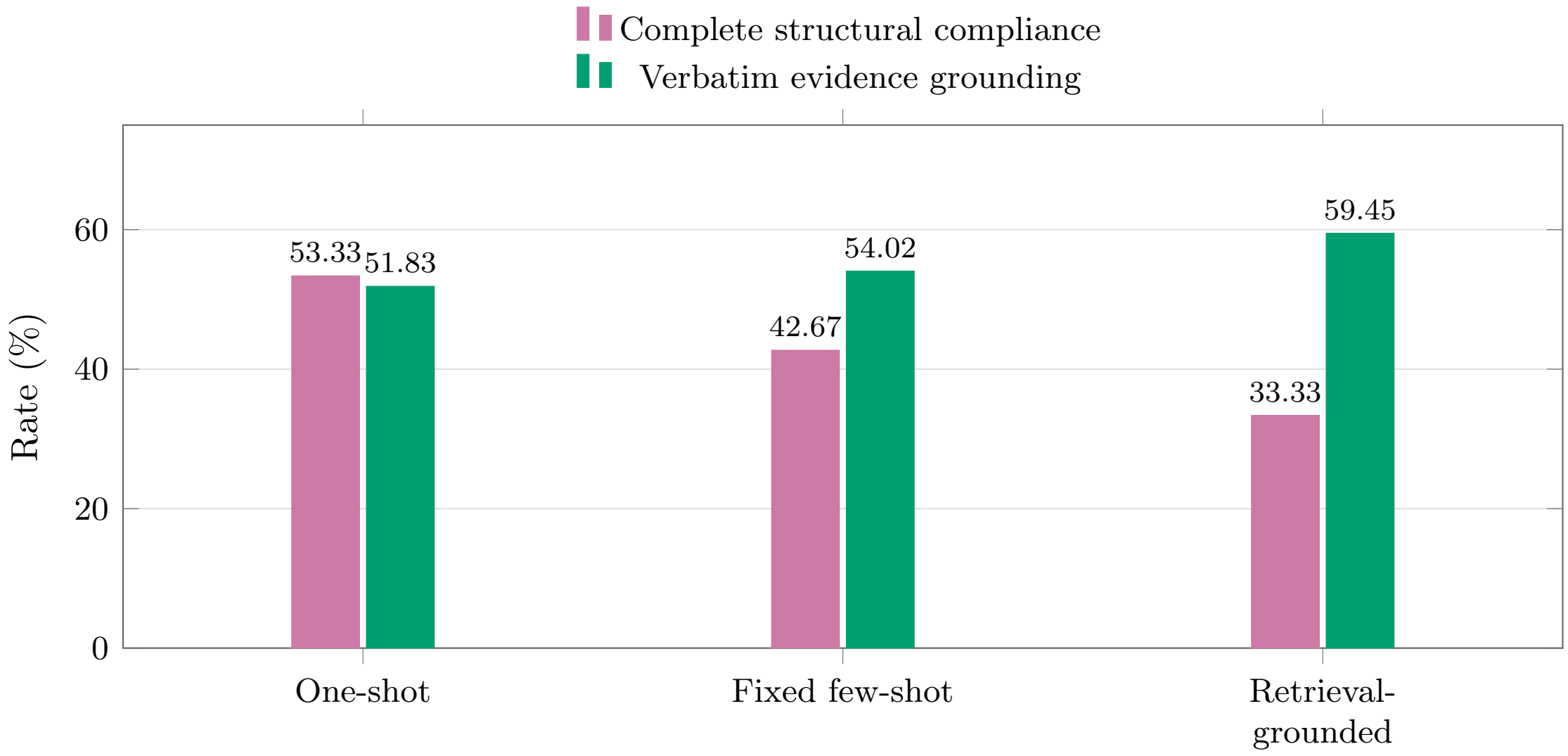

Figure~\ref{fig:structural_grounding} illustrates the contrasting patterns across the demonstration-based conditions, with one-shot prompting achieving the highest complete structural compliance and retrieval-grounded prompting achieving the highest proportion of verbatim evidence spans.

\subsection{Relation-Level Extraction Performance}

Across the four prompting conditions, SmolLM2 generated 1,804 relations, of which 736 (40.8\%) were structurally malformed and 1,068 satisfied the structural requirements for relation evaluation. Malformed relations remained false positives rather than being removed or repaired, ensuring that the reported scores reflected the generated outputs and their adherence to the specified schema.

Table~\ref{tab:relation_results} summarizes relation extraction against the reconciled reference containing 467 directed relations per condition. Finalized one-to-one matching identified 82 true positives across the four evaluations, together with 1,722 false positives and 1,786 false negatives, yielding aggregate precision of 4.55\%, recall of 4.39\%, and F1 of 4.47\%.

\begin{table}[htbp]
\caption{Directed relation extraction against the reconciled reference containing 467 relations per condition. Precision, recall, and F1 are percentages, while the aggregate row pools counts across all four conditions.}
\label{tab:relation_results}
\centering
\small
\setlength{\tabcolsep}{4pt}
\begin{tabular}{lrrrrrrr}
\toprule
\textbf{Condition} & \textbf{Pred.} & \textbf{Ref.} & \textbf{TP} & \textbf{FP} & \textbf{FN} & \textbf{P} & \textbf{R} \\
\midrule
Zero-shot          & 129  & 467  & 0  & 129  & 467  & 0.00 & 0.00 \\
One-shot           & 281  & 467  & 26 & 255  & 441  & 9.25 & 5.57 \\
Fixed few-shot     & 743  & 467  & 31 & 712  & 436  & 4.17 & 6.64 \\
Retrieval-grounded & 651  & 467  & 25 & 626  & 442  & 3.84 & 5.35 \\
\midrule
Aggregate          & 1804 & 1868 & 82 & 1722 & 1786 & 4.55 & 4.39 \\
\bottomrule
\end{tabular}

\medskip
\begin{tabular}{lrl}
\toprule
\textbf{Condition} & \textbf{F1 (\%)} & \textbf{95\% CI} \\
\midrule
Zero-shot          & 0.00          & 0.00--0.00 \\
One-shot           & \textbf{6.95} & 4.49--9.63 \\
Fixed few-shot     & 5.12          & 3.34--7.03 \\
Retrieval-grounded & 4.47          & 2.53--6.60 \\
\bottomrule
\end{tabular}
\end{table}

One-shot prompting achieved the highest observed relation F1 of 6.95\%, recovering 26 reference relations from 281 predictions, whereas fixed few-shot prompting recovered 31 reference relations from 743 predictions and achieved an F1 of 5.12\%. Retrieval-grounded prompting recovered 25 reference relations from 651 predictions and achieved an F1 of 4.47\%, while zero-shot prompting recovered no reference relations.

Paired narrative-level bootstrap analysis yielded relation F1 estimates of 6.95\% (95\% CI: 4.49--9.63) for one-shot prompting, 5.12\% (3.34--7.03) for fixed few-shot prompting, and 4.47\% (2.53--6.60) for retrieval-grounded prompting. The paired retrieval-grounded-minus-fixed-few-shot difference was $-0.65$ percentage points (95\% CI: $-3.39$ to $2.12$), providing insufficient evidence of a consistent difference between these two conditions.

Zero-shot prompting produced an empirical bootstrap interval of 0.00--0.00\% because no true-positive relation occurred in any evaluated narrative; this degenerate interval describes the observed sample and should not be interpreted as certainty about performance on unseen narratives.

Malformed outputs frequently used labels outside the permitted vocabulary, including \texttt{RESULT\_IN} in 349 instances, \texttt{RESULT\_FROM} in 120 instances, and \texttt{RESULT\_OF} in 68 instances, while another 180 instances had empty relation labels. These outputs were retained as errors because the evaluation required the specified relation type and direction alongside compatible source and target concepts.

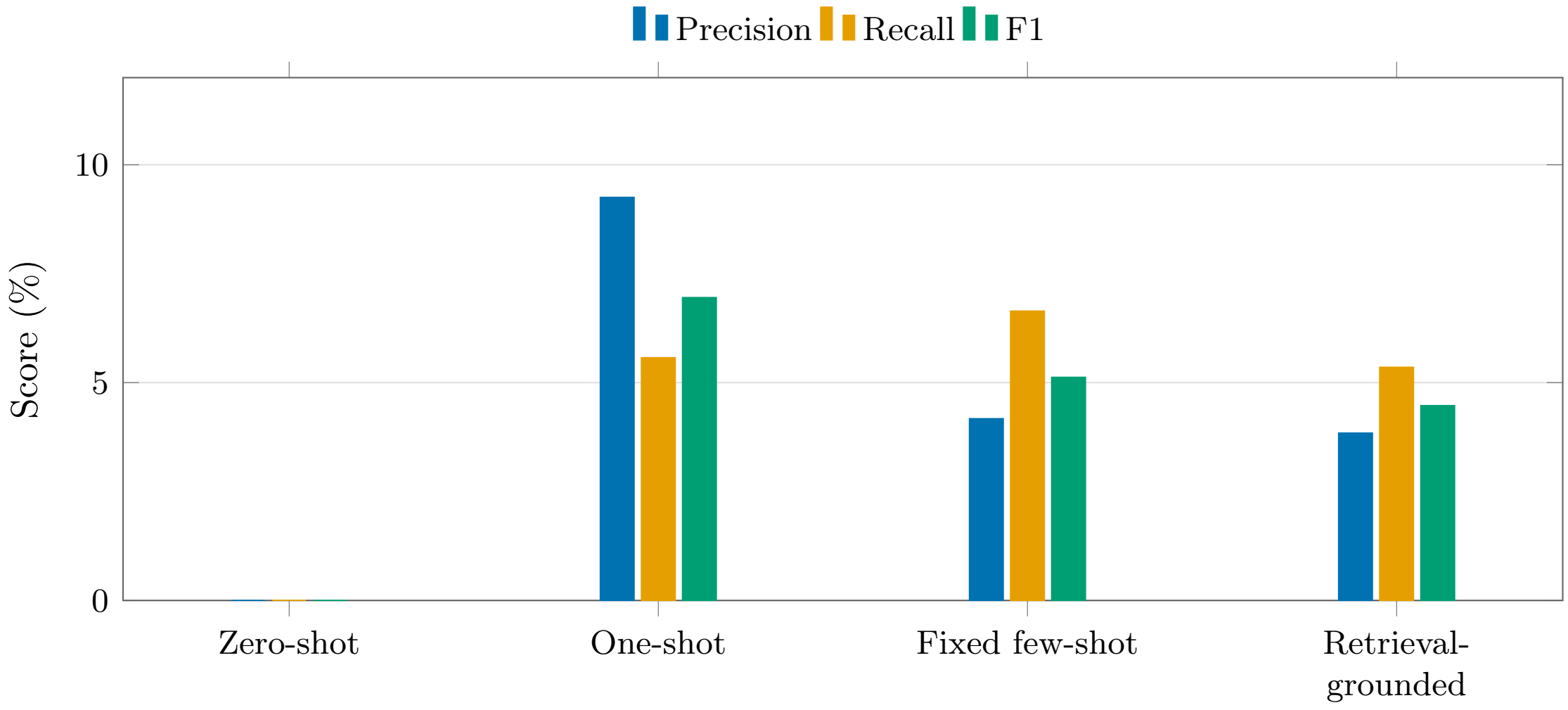
\begin{figure}[htbp]
\centering
\begin{tikzpicture}
\begin{axis}[
    makeaxis,
    ybar,
    bar width=9pt,
    ymin=0,
    ymax=12,
    ylabel={Score (\%)},
    symbolic x coords={Zero,One,Fixed,Retrieved},
    xtick=data,
    xticklabels={
        Zero-shot,
        One-shot,
        Fixed few-shot,
        Retrieval-grounded
    },
    x tick label style={
        font=\footnotesize,
        text width=2.3cm,
        align=center
    },
    legend style={
        at={(0.5,1.03)},
        anchor=south,
        legend columns=3,
        draw=none,
        fill=none,
        font=\footnotesize
    }
]

\addplot[
    fill=makeblue,
    draw=makeblue
] coordinates {
    (Zero,0.00)
    (One,9.25)
    (Fixed,4.17)
    (Retrieved,3.84)
};

\addplot[
    fill=makeorange,
    draw=makeorange
] coordinates {
    (Zero,0.00)
    (One,5.57)
    (Fixed,6.64)
    (Retrieved,5.35)
};

\addplot[
    fill=makegreen,
    draw=makegreen
] coordinates {
    (Zero,0.00)
    (One,6.95)
    (Fixed,5.12)
    (Retrieved,4.47)
};

\legend{Precision,Recall,F1}

\end{axis}
\end{tikzpicture}
\caption{Directed relation precision, recall, and F1 across
the four prompting conditions, evaluated against 467 reference
relations in 300 narratives. Correct matches require compatible
source and target concepts, the specified relation type,
and the correct direction. All values are percentages.}
\label{fig:relation_performance}
\end{figure}

Figure~\ref{fig:relation_performance} compares relation precision, recall, and F1 across the four prompting conditions, illustrating that the higher number of generated relations under fixed few-shot and retrieval-grounded prompting did not translate into higher F1 than one-shot prompting.

\subsection{Concept--Relation Performance Gap}

A pronounced performance gap was observed between concept identification and directed relation extraction, with the highest concept micro-F1 of 32.27\% achieved under retrieval-grounded prompting and the highest relation F1 of 6.95\% achieved under one-shot prompting. Within individual conditions, one-shot prompting achieved concept and relation F1 scores of 30.75\% and 6.95\%, fixed few-shot prompting achieved 31.52\% and 5.12\%, and retrieval-grounded prompting achieved 32.27\% and 4.47\%, respectively.

The strongest prompting condition therefore differed between concept and relation extraction, while the greater number of relations generated under the four-demonstration conditions did not produce higher relation F1 than one-shot prompting. Correct relation extraction required compatible source and target concepts, the specified relation type, and the correct direction, meaning that individually plausible concepts could still form an incorrect relation because of an unsuitable endpoint pairing, a mismatched label, a reversed direction, or an omitted intermediate event.

\subsection{Exploratory Comparison with Qwen3-8B}
\label{sec:qwen-results}

Table~\ref{tab:qwen-comparison} compares Qwen3-8B and SmolLM2 on the same 35 narratives against the reconciled subset reference containing 432 concepts and 77 directed relations. SmolLM2 scores were recomputed from its existing outputs for these narratives, ensuring that comparisons used matched cases rather than the full-benchmark SmolLM2 scores. Qwen3-8B produced parseable JSON objects in all 140 responses, and none reached the generation limit.

Qwen3-8B achieved higher observed concept micro-F1 and complete structural compliance under all four prompting conditions, together with higher relation F1 and verbatim evidence grounding under all three demonstration-based conditions. Its highest concept micro-F1 and relation F1 were 51.37\% and 34.62\%, respectively, both under one-shot prompting, whereas its highest complete structural compliance of 85.71\% and highest demonstration-based grounding rate of 75.47\% occurred under retrieval-grounded prompting.

\begin{table}[htbp]
\caption{Exploratory model comparison on the same 35 narratives against the reconciled reference containing 432 concepts and 77 directed relations. Concept micro-F1, relation F1, complete structural compliance, and verbatim evidence grounding are percentages.}
\label{tab:qwen-comparison}
\centering
\small
\setlength{\tabcolsep}{3pt}
\begin{tabular}{llrrrr}
\toprule
\textbf{Model}
& \textbf{Condition}
& \shortstack{\textbf{Concept}\\\textbf{F1}}
& \shortstack{\textbf{Relation}\\\textbf{F1}}
& \textbf{Structure}
& \textbf{Grounding} \\
\midrule
SmolLM2 & Zero-shot          & 3.43  & 0.00  & 5.71  & -- \\
SmolLM2 & One-shot           & 35.68 & 5.36  & 60.00 & 56.50 \\
SmolLM2 & Fixed few-shot     & 36.07 & 7.32  & 34.29 & 58.66 \\
SmolLM2 & Retrieval-grounded & 36.94 & 1.25  & 42.86 & 62.37 \\
\midrule
Qwen3-8B & Zero-shot          & 46.29 & 0.00  & 34.29 & 78.91 \\
Qwen3-8B & One-shot           & 51.37 & 34.62 & 74.29 & 70.37 \\
Qwen3-8B & Fixed few-shot     & 49.86 & 24.28 & 80.00 & 74.91 \\
Qwen3-8B & Retrieval-grounded & 48.63 & 27.22 & 85.71 & 75.47 \\
\bottomrule
\end{tabular}
\end{table}

Under zero-shot prompting, Qwen3-8B achieved a verbatim-grounding rate of 78.91\%, whereas SmolLM2 produced no usable evidence spans and therefore had an undefined grounding rate. These measurements establish the presence of generated evidence spans in the source narratives rather than whether those spans support the associated concepts.

Table~\ref{tab:qwen-paired-differences} reports paired Qwen3-8B-minus-SmolLM2 differences in concept micro-F1 and relation F1, with intervals calculated from 10,000 paired narrative-level bootstrap resamples. Concept-F1 differences favored Qwen3-8B under all four prompting conditions, and each individual 95\% CI excluded zero.

\begin{table}[htbp]
\caption{Paired Qwen3-8B-minus-SmolLM2 differences on the same 35 narratives. Differences and confidence limits are expressed in percentage points; intervals are unadjusted for multiple comparisons and conditional on the finalized reference and matching decisions.}
\label{tab:qwen-paired-differences}
\centering
\small
\setlength{\tabcolsep}{4pt}
\begin{tabular}{lrlrl}
\toprule
\textbf{Condition}
& \shortstack{\textbf{Concept}\\\textbf{difference}}
& \textbf{95\% CI}
& \shortstack{\textbf{Relation}\\\textbf{difference}}
& \textbf{95\% CI} \\
\midrule
Zero-shot
& 42.86 & 37.71--48.36
& 0.00 & 0.00--0.00 \\
One-shot
& 15.69 & 8.66--22.78
& 29.26 & 14.88--43.19 \\
Fixed few-shot
& 13.79 & 8.34--19.13
& 16.96 & 5.22--27.96 \\
Retrieval-grounded
& 11.69 & 6.22--16.89
& 25.97 & 15.49--36.70 \\
\bottomrule
\end{tabular}
\end{table}

For relation extraction, paired differences favored Qwen3-8B under all three demonstration-based conditions, with individual 95\% CIs excluding zero, while both models recovered no correct relations under zero-shot prompting. The zero-shot relation-difference interval of 0.00--0.00 percentage points reflects the absence of observed true positives for either model and should not be interpreted as evidence of equivalent performance on unseen narratives.

The completed supplementary matching exercise incorporated prior adjudicated decisions, applicable reused matches, and 374 additional review tasks comprising 295 concept tasks and 79 relation tasks. All additional tasks were resolved before final scoring, but the retained records do not establish uniformly independent pre-adjudication judgments across the expanded exercise; consequently, the earlier matching-agreement coefficients are not reported as agreement estimates for this reconciled evaluation.

These results support higher concept extraction performance for Qwen3-8B within the evaluated subset and higher relation F1 under demonstration-based prompting, although the individual intervals were not adjusted for multiple comparisons. Selection of the 35 narratives by processing order under computational constraints limits generalization, while differences in training, architecture, and parameter count prevent attribution of the observed differences specifically to model size.

\subsection{Summary of Results}

The main 300-narrative experiment showed that annotated demonstrations substantially improved SmolLM2 concept extraction relative to zero-shot prompting, with the largest increase occurring after the introduction of one annotated example. 
Retrieval-grounded prompting achieved the highest observed concept micro-F1 (32.27\%) and macro-F1 (20.76\%). For both metrics, the 95\% confidence interval for the paired difference relative to fixed few-shot prompting included zero. 

Concept accuracy, structural compliance, and verbatim grounding did not improve in parallel, with retrieval-grounded prompting achieving the highest verbatim-grounding rate of 59.45\% and one-shot prompting achieving the highest complete structural compliance of 53.33\%. Category-level performance varied substantially, and concept micro-F1 declined across increasing narrative-length strata under every prompting condition.

Directed relation extraction remained the principal bottleneck, with the highest relation F1 of 6.95\% achieved under one-shot prompting and an aggregate relation F1 of 4.47\% across the four conditions, while 40.8\% of generated relations were structurally malformed. These findings show that improvements in concept identification did not ensure accurate reconstruction of directed relationships.

The matched 35-narrative comparison showed higher concept micro-F1 and complete structural compliance for Qwen3-8B across all four prompting conditions, together with higher relation F1 and verbatim grounding under the three demonstration-based conditions. Individual paired confidence intervals excluded zero for all concept-F1 differences and all three demonstration-based relation-F1 differences, although these exploratory findings remain conditional on the selected subset and finalized matching decisions.

\section{Discussion}
\label{sec:discussion}

The results show that annotated demonstrations substantially improved SmolLM2 concept extraction, but their benefits differed across semantic accuracy, source traceability, structural compliance, and directed relation correctness. Retrieval-grounded prompting achieved the highest observed concept micro-F1, macro-F1, and verbatim-grounding rate, whereas one-shot prompting achieved the highest complete structural compliance and relation F1.  The 95\% confidence intervals for the paired differences in concept micro-F1 and macro-F1 both included zero. The higher point estimates therefore do not establish an advantage for retrieval-grounded prompting over fixed few-shot prompting on either metric. 

\subsection{Effect of In-Context Demonstrations on Concept Extraction}

Concept micro-F1 increased from 3.72\% under zero-shot prompting to 30.75\% under one-shot prompting. The number of responses that could be parsed as JSON also increased from 71 to 256 out of 300 attempts. Because unparseable responses contributed no concept predictions, the F1 comparison reflects both the model’s ability to return a parseable response and its ability to identify concepts within that response.  Demonstrations may clarify category boundaries, extraction granularity, and the expected relationship between narrative content and structured output, although the experiment did not isolate the contribution of each component.

The smaller gains under fixed few-shot and retrieval-grounded prompting suggest diminishing improvements from additional demonstrations within the evaluated configuration. Retrieval-grounded prompting achieved a micro-F1 of 32.27\%, compared with 31.52\% under fixed few-shot prompting, but the paired difference of 0.75 percentage points had a 95\% CI of $-0.92$ to $2.40$. The corresponding macro-F1 difference was also uncertain at 0.02 percentage points (95\% CI: $-1.36$ to $1.39$), indicating that neither aggregate measure established a consistent advantage for dynamic demonstration selection.

These findings complement aviation research on retrieval-supported extraction and reasoning \cite{Ren2025,Cheng2026} . On this benchmark, selecting demonstrations by narrative similarity yielded no clear advantage over the fixed demonstration set in either concept micro-F1 or macro-F1 . An example may resemble the target report without illustrating its less frequent categories, evidence patterns, or relation configurations, motivating selection methods that consider schema coverage alongside narrative similarity.

\subsection{Category-Specific Extraction Behavior}

The comparatively strong extraction of phases of flight and consequences suggests that recognizable operational expressions and explicitly described outcomes may be easier for the evaluated model to recover than concepts requiring finer distinctions about their investigative role. However, linguistic explicitness was not measured directly, and the observed category differences should not be attributed to this explanation alone.

Category imbalance and sparse predictions also affect interpretation, particularly when high precision accompanies very low recall. For example, the 100\% technical-factor precision under fixed few-shot prompting arose from one correct prediction against 154 reference instances, while one-shot prompting correctly recovered only one of 394 environmental-factor instances. These results illustrate why precision alone would overstate category coverage and why micro-F1, macro-F1, and category-specific measures should be considered together.

The schema further requires the model to distinguish a condition's textual presence from its conceptual role, because a weather statement does not automatically constitute an environmental factor, an aircraft component does not automatically constitute a technical factor, and a human action does not automatically constitute a human factor. Errors can therefore arise from incorrect interpretation even when the extracted wording appears in the narrative.

The demonstration audit provides a possible explanation for some category-specific differences, although it does not establish causation. Retrieved sets contained an average of 2.03 environmental-factor concepts per prompt, compared with one in the fixed set, alongside an increase in environmental-factor F1 from 0.51\% to 12.64\%; conversely, human-factor coverage averaged 2.15 concepts under retrieval rather than four, alongside a decrease in human-factor F1 from 16.00\% to 5.38\%.

Selection was also concentrated, with DEV\_07 and DEV\_03 occupying 447 of the 1,200 retrieval slots, whereas DEV\_10, the only demonstration containing technical-factor and contributing-factor concepts, appeared in 26 prompts. Dynamic retrieval therefore did not guarantee balanced schema coverage, motivating controlled evaluation of category-aware and diversity-aware demonstration selection.

\subsection{Association Between Narrative Length and Extraction Performance}

The consistent decline in concept micro-F1 across increasing narrative-length strata suggests that more extensive reports posed greater extraction challenges under every prompting condition. Under retrieval-grounded prompting, recall decreased from 32.94\% for short narratives to 16.06\% for long narratives, indicating that the model recovered a smaller proportion of the annotated concepts as narrative length increased.

Longer reports may contain more contextual details, interacting events, and distinctions between background information and safety-relevant findings, but the observational analysis cannot separate narrative length from concept density, event complexity, or causal structure. The fixed generation budget may also interact with the amount of information requiring extraction, although generation-limit hits alone do not establish which concepts were omitted or why.

The strongest observed condition differed across strata, with one-shot prompting performing best for short narratives, fixed few-shot prompting for medium narratives, and retrieval-grounded prompting for long narratives. These descriptive differences motivate adaptive extraction strategies, but they do not establish statistically reliable superiority within individual strata.

\subsection{Evidence Grounding and Structural Reliability}

The higher verbatim-grounding rate and lower complete structural compliance under retrieval-grounded prompting demonstrate that traceability and output reliability did not improve together. Relative to fixed few-shot prompting, retrieval increased verbatim grounding by 5.42 percentage points (95\% CI: 1.89--8.85) but reduced complete compliance by 9.33 percentage points (95\% CI: $-17.00$ to $-1.33$), supporting the observed trade-off within this benchmark.

Verbatim grounding establishes only that a generated evidence span occurs in the source narrative, whereas determining whether it supports the associated concept, category assignment, or relation requires an additional relevance assessment. This distinction is consistent with work separating generated-answer correctness from attribution quality \cite{Gao2023ALCE} and prevents the higher retrieval-grounding rate from being interpreted as evidence of more faithful reasoning.

Zero-shot prompting produced no usable evidence spans, making its grounding rate undefined rather than zero, while its high frequency of generation-limit hits suggests difficulty maintaining the requested output format without demonstrations. Nevertheless, the lower complete compliance under the four-demonstration conditions relative to one-shot prompting shows that additional examples did not consistently improve schema adherence.

Greater contextual complexity is one possible explanation, but demonstration number, content, and prompt length changed together, preventing attribution of the observed decline to any single factor. For practical use, quoted evidence may improve auditability, but source presence, semantic interpretation, and category assignment still require separate verification.

\subsection{Relation Extraction as the Principal Bottleneck}

The gap between the highest concept micro-F1 of 32.27\% and the highest relation F1 of 6.95\% identifies directed relation extraction as the principal limitation of the evaluated SmolLM2 configurations. Correct relations require compatible endpoint meanings, the specified relation label, and the correct direction, so recognizing plausible concepts does not ensure accurate reconstruction of their relationship.

Strict endpoint matching also distinguishes a direct annotated relation from a broader causal association, because matching an outcome does not compensate for an upstream source or an omitted intermediate event. Similarly, substituting \texttt{CONTRIBUTES\_TO} for \texttt{RESULTS\_IN} changes the relation being evaluated even when the endpoints are otherwise compatible.

The 40.8\% malformed-relation rate indicates an additional structural limitation, including frequent production of labels outside the permitted vocabulary. A normalization or constrained-generation stage could reduce such errors, but any improvement should be evaluated separately from semantic correctness because valid labels can still connect incorrect endpoints.

Additional demonstrations increased relation generation without improving F1 over one-shot prompting, as the modest increase in true positives under fixed few-shot prompting was accompanied by substantially more false positives. These findings caution against treating a larger generated graph as a more complete or reliable representation of an investigation.

For aviation knowledge-graph construction, relation quality therefore requires explicit assessment before generated edges support retrieval, causal analysis, or decision support \cite{Chen2024CausationKG,Chen2026ARCEM,Cheng2026}. The related decomposed study evaluates a complementary setting with supplied reference endpoints \cite{NanyongaDecomposed}, whereas the present task combines concept generation with relation reconstruction.

\subsection{Interpretation of the Matched Qwen3-8B Comparison}

The matched comparison indicates that the severity of the extraction limitations differed between the evaluated models, with Qwen3-8B achieving higher concept micro-F1 and complete structural compliance under all four conditions and higher relation F1 under all demonstration-based conditions. Its strongest concept and relation F1 scores, 51.37\% and 34.62\%, respectively, both occurred under one-shot prompting.

Individual paired confidence intervals excluded zero for all concept-F1 differences and all three demonstration-based relation-F1 differences, supporting an advantage for Qwen3-8B within the evaluated subset, conditional on the reconciled reference and matching decisions. Nevertheless, its best relation score still indicates substantial remaining errors, while the zero relation F1 achieved by both models under zero-shot prompting shows that stronger concept extraction alone did not resolve relation reconstruction.

The comparison remains exploratory because the 35 narratives were selected by processing order under computational constraints and the intervals were not adjusted for multiple comparisons. Differences in training, architecture, and parameter count also prevent attribution of the observed advantage specifically to model size, so the results motivate broader matched evaluation rather than a general scaling conclusion.

\subsection{Implications for Aviation Safety Knowledge Extraction}

The findings support selecting extraction configurations according to several requirements rather than a single score, because the condition with the highest observed concept performance and verbatim grounding did not achieve the strongest structural compliance or relation F1. Demonstrations substantially improved the compact model without task-specific fine-tuning, but prompting alone did not resolve low recall, uneven category coverage, or unreliable directed relations.

The generated outputs are therefore best treated as candidates for analyst review, with concepts, evidence, and relations verified separately before acceptance into a safety knowledge base. Evidence-linked predictions may help reviewers locate relevant passages, but the study did not measure analyst workload or downstream safety outcomes, so operational benefits remain to be established.

\subsection{Interpretation of Reviewer Agreement}

Final reconciliation establishes the decisions used for scoring, whereas independent agreement characterizes consistency before discussion or adjudication, making confirmation of the final reference distinct from an independent agreement measurement. The category-presence diagnostics for 232 eligible narratives concern whether a category was populated and do not establish agreement on concept identity, completeness, evidence relevance, or complete directed relations.

Prevalence also affects interpretation, as illustrated by high unanimity but low Fleiss' $\kappa$ for event and contributing-factor presence. Reporting unanimity alongside chance-corrected agreement exposes these effects and avoids interpreting agreement on widespread presence or absence as agreement on individual annotations.

The expanded semantic-matching records contain adjudicated decisions, reused decisions, supplied reviewer fields, and subsequent corrections, but do not establish uniformly independent pre-adjudication judgments across the completed exercise. Consequently, the reported bootstrap intervals quantify narrative-level variation conditional on the finalized decisions rather than annotation or adjudication uncertainty.

\subsection{Limitations and Future Work}

The main evaluation used one compact model and 300 narratives from a single reporting authority and temporal range, while the supplementary comparison covered a nonrandom subset of 35 narratives. Evaluation across the complete benchmark, additional model families, reporting authorities, and time periods is needed to assess generalization and determine whether the observed prompting trade-offs persist.

The reconciled reference and finalized matching decisions define the scope of the semantic results, while limited representation of some categories, particularly the nine contributing-factor instances, restricts category-specific conclusions. Versioned annotations, reproducible scoring, and independently retained reviewer submissions would support more transparent comparisons and allow future uncertainty analyses to incorporate annotation variability.

The retrieval experiment used a small pool of 12 demonstrations and did not independently manipulate category coverage, selection diversity, example order, or prompt length. Larger pools and controlled ablations could clarify which demonstration properties improve extraction, while segmentation and hierarchical methods could test whether performance on information-dense narratives can be improved.

Verbatim grounding does not assess concept--evidence support and rejects paraphrases that are absent as exact substrings, so future evaluation should combine source-presence checks with explicit judgments of evidential relevance. Such judgments should remain distinct from concept matching because a correct concept may still be paired with an irrelevant quotation.

Finally, unconstrained generation and the fixed output budget may have contributed to structural errors and incomplete extraction, motivating separate evaluations of output-budget changes, schema-constrained decoding, relation-label enforcement, and staged relation verification. These comparisons should distinguish improvements in formatting from improvements in semantic extraction, while prospective analyst studies should assess whether verified outputs reduce review effort without introducing additional errors.

\section{Conclusions}
\label{sec:conclusions}

This study evaluated SmolLM2-1.7B-Instruct under zero-shot, one-shot, fixed few-shot, and retrieval-grounded prompting on 300 NTSB narratives, using a reconciled reference containing 3,763 concepts across eight categories and 467 directed relations of two types. Separate assessments of concept accuracy, verbatim evidence grounding, structural compliance, and relation extraction showed that no prompting strategy performed best across all four dimensions.

Annotated demonstrations substantially improved concept extraction, with micro-F1 increasing from 3.72\% under zero-shot prompting to 30.75\% under one-shot prompting, 31.52\% under fixed few-shot prompting, and 32.27\% under retrieval-grounded prompting. Although retrieval achieved the highest observed micro- and macro-F1, the 95\% confidence intervals for the paired differences from fixed few-shot prompting included zero, providing insufficient evidence of an aggregate concept-extraction advantage for either four-demonstration strategy. 

Retrieval-grounded prompting achieved a higher verbatim-grounding rate than fixed few-shot prompting, with a paired difference of 5.42 percentage points (95\% CI: 1.89--8.85), but lower complete structural compliance, with a difference of $-9.33$ percentage points (95\% CI: $-17.00$ to $-1.33$). These findings demonstrate a trade-off between source traceability and schema compliance, while verbatim presence alone does not establish that the quoted text supports the associated concept.

Directed relation extraction remained the principal bottleneck, with the highest relation F1 of 6.95\% achieved under one-shot prompting and 40.8\% of generated relations structurally malformed across the four conditions. Concept extraction also varied substantially across categories and declined across increasing narrative-length strata, indicating that aggregate performance concealed important differences in extraction coverage.

The exploratory comparison on 35 matched narratives showed higher concept micro-F1 and complete structural compliance for Qwen3-8B under all four conditions, together with higher relation F1 under all three demonstration-based conditions. Its highest concept and relation F1 scores were 51.37\% and 34.62\%, respectively, both under one-shot prompting, and individual paired confidence intervals excluded zero for all concept-F1 differences and all demonstration-based relation-F1 differences. These findings remain conditional on the selected subset and finalized matching decisions, while the nonrandom subset and differences between model configurations restrict broader conclusions.

The results support evaluating extraction performance across multiple dimensions and treating generated concepts and relations as candidates for human review. Reviewers should verify each concept, its quoted evidence, and each proposed relation before downstream use. Future work should extend matched model comparisons, investigate demonstration coverage and constrained generation, and assess evidential relevance and analyst utility alongside extraction accuracy.
\section*{Author Contributions}

Conceptualization, A.N.; methodology, A.N. and A.H.;
software, A.N.; validation, A.N., G.W. and U.T.;
formal analysis, A.N.; investigation, A.N.;
data curation, A.N.; writing---original draft
preparation, A.N., M.T.Y.; writing---review and
editing, A.N., M.T.Y., U.T., G.W. and A.H.;
visualization, A.N.; supervision, G.W. and A.H.
All authors have read and approved the manuscript.

\section*{Funding}

This research was funded by the Deanship of Scientific
Research (DSR) at King Abdulaziz University, Jeddah,
Saudi Arabia, under grant no. IPP: 207-135-2025.

\section*{Data Availability}

The aviation accident data analyzed in this study were
obtained from publicly available National Transportation
Safety Board (NTSB) records. The annotated benchmark,
development examples, model outputs, and evaluation
materials generated during this study will be made
available in a public repository upon journal publication,
subject to the data-sharing conditions applicable to
the original NTSB records.

\section*{Acknowledgments}

The authors acknowledge with thanks DSR for technical
and financial support.

\section*{Conflicts of Interest}

The authors declare no conflicts of interest.

\section*{Abbreviations}{%
The following abbreviations are used in this manuscript:\\

\noindent
\begin{tabular}{@{}ll}
AI & Artificial Intelligence\\
ARCEM & Aviation Reports Causality Extraction Model\\
BERT & Bidirectional Encoder Representations from Transformers\\
FP & False Positive\\
FN & False Negative\\
GPU & Graphics Processing Unit\\
JSON & JavaScript Object Notation\\
LLM & Large Language Model\\
NLP & Natural Language Processing\\
NTSB & National Transportation Safety Board\\
NF4 & NormalFloat4\\
RAG & Retrieval-Augmented Generation\\
RQ & Research Question\\
TP & True Positive\\
\end{tabular}
}


\end{document}